%% file: iclr2027_conference.tex
\documentclass{article} 
\usepackage{iclr2027_conference,times}

\input{math_commands.tex}

\usepackage{hyperref}
\usepackage{url}

\usepackage{mdframed}
\usepackage{array}
\usepackage{longtable}
\usepackage[most]{tcolorbox}
\usepackage[font=small]{caption}
\usepackage{amsmath}
\usepackage{graphicx}
\usepackage{multirow}
\usepackage{booktabs}
\usepackage{multirow}
\usepackage{enumitem}
\usepackage{wrapfig}
\usepackage{hyperref}
\usepackage{subcaption}

\newtcolorbox{llmverbatim}[1]{enhanced, breakable,
   colback=gray!4!white, colframe=gray!50!black, boxrule=0.6pt, arc=1.5mm,
   left=10pt, right=10pt, top=8pt, bottom=8pt,
   fonttitle=\bfseries\small, title={Verbatim reasoning trace -- #1}}

\newtcolorbox{llmsummary}[1]{enhanced, breakable,
 colback=gray!4!white, colframe=gray!50!black, boxrule=0.6pt, arc=1.5mm,
   left=10pt, right=10pt, top=8pt, bottom=8pt,
   fonttitle=\bfseries\small, title={API-summarised reasoning trace -- #1}}

\newtcolorbox{authorprompt}{enhanced, breakable,
   colback=white, colframe=gray!70!black, boxrule=0.6pt, arc=1.5mm,
   left=10pt, right=10pt, top=8pt, bottom=8pt, dashed,
   fonttitle=\bfseries\small, title={Author-written prompt fragment}}

\newcounter{result}
\renewcommand{\theresult}{\arabic{result}}

\newenvironment{boxedresult}
  {%
    \refstepcounter{result}%
    \begin{mdframed}%
    \noindent\textbf{Result~\theresult.}\quad
  }
  {%
    \end{mdframed}%
  }

\newlength{\DutchBookBoxHeight}
\newlength{\DutchBookFirstPartHeight}
\title{From Prompts to Posteriors: Can LLM Agents Perform Bayesian Inference in Context?}

\author{%
  \centerline{Linus Filbry$^{1}$ \qquad
  Luckeciano C. Melo$^{1}$ \qquad
  Yarin Gal$^{1}$} \\[0.5em]
  \centerline{{\normalsize $^{1}$OATML, University of Oxford}}\\[0.6em]
  \centerline{{\small\itshape
  linus.filbry@mailbox.org \quad
  luckeciano.carvalhomelo@wolfson.ox.ac.uk}}
}

\iclrfinalcopy 
\begin{document}

\maketitle

\begin{abstract}
Large language models (LLMs) are increasingly applied across the scientific discovery pipeline, from synthesizing literature to generating hypotheses to informing experimental design. A unifying principle behind these tasks is the need to continuously update beliefs as evidence accumulates, for which Bayesian inference provides a natural framework. However, existing work assumes that LLMs cannot carry out principled Bayesian belief updating in context, instead relying on simplifications and heuristics. To challenge this assumption, we present a principled evaluation of Bayesian reasoning capabilities in LLMs, analyzing both the model's internal consistency as well as its ability to recover an externally determined posterior. Specifically, we build an evaluation framework spanning qualitative epistemological examples to probe probabilistic coherence, quantitative benchmarks of Bayesian inference capabilities, and analyses of the impact of non-Bayesian updates on downstream task performance. Across different scenarios, we find that LLMs generally violate Bayes' rule when updating their beliefs in context. However, when equipped with modern reasoning abilities and code execution capabilities, their performance drastically improves, enabling even the close approximation of analytically intractable posteriors. Our evaluation shows that current LLMs with access to appropriate tools could already be used to make complex belief updates in scientific discovery, enabling rapid acceleration of the efficiency and scale of AI-powered scientific discovery.
\end{abstract}

\section{Introduction}\label{sec:introduction}
LLMs are rapidly moving from passive tools for retrieving and summarizing scientific information toward active components of the scientific workflow. Recent systems use LLMs to generate hypotheses, synthesize literature, write and execute code, design experiments, interpret results, and even assemble full research papers \citep{lu2024aiscientist, aiscientist_survey1, aiscientist_survey2, aiscientist_survey3}. This trend has motivated the broader vision of ``AI scientists'': autonomous or semi-autonomous agents that participate in the iterative production of scientific knowledge rather than merely assisting with isolated subtasks.

A central requirement for such agents is the ability to \textit{rationally} revise beliefs in light of evidence, iteratively reallocating probability to hypotheses which best explain current experimental results \citep{bayesian_scientific_reasoning, bayesian_scientific_reasoning2}. However, current AI-scientist systems typically take a significant shortcut, which strongly limits their potential performance: They evaluate hypotheses through task-specific heuristics or experimental scores, rather than by maintaining and updating probabilistic beliefs over competing hypotheses \citep{song2025evaluating, aiscientist_survey3}.

Bayesian inference provides a natural framework to satisfy this requirement. It enables both explicit incorporation of the extensive prior knowledge LLMs possess and information-theoretically optimal belief updates when presented with new evidence \citep{bayesian_decisiontheoretic}. Using non-Bayesian belief updating rules risks introducing logical incoherence and biases into the AI scientist's scientific reasoning \citep{bayesian_scientific_reasoning}. This allows overreacting to noisy experimental anomalies or prematurely discarding valid hypotheses \citep{jaynes2003probability}. An ideal AI scientist must therefore be capable of performing Bayesian inference to correctly update its beliefs when presented with complex evidence. This requirement exposes a critical research question for the development of LLM-driven research agents: \textit{To what extent can LLMs perform or approximate Bayesian inference in context?}

\begin{figure}[t]
    \centering
    \includegraphics[width=\linewidth]{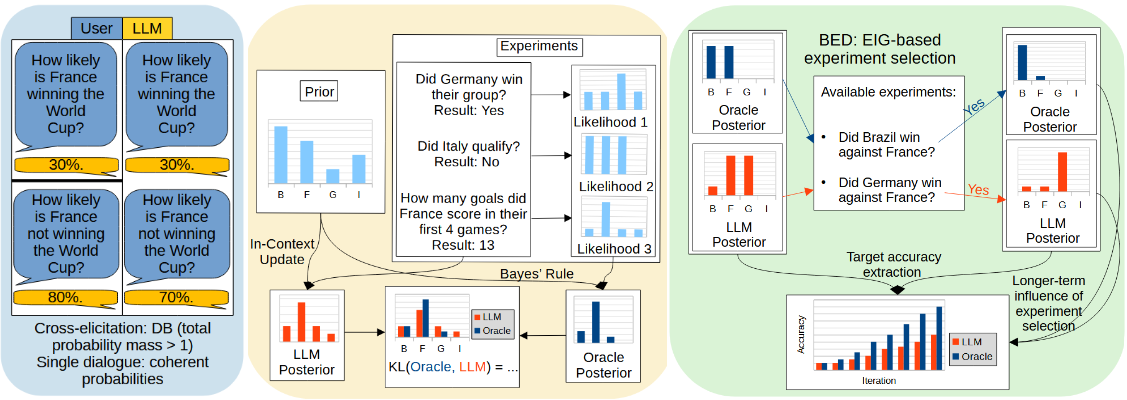}
    \caption{\textbf{Overview of our evaluation framework for In-Context Bayesian Inference} We present our three axes of evaluation, using example task: "Who will win the World Cup - Brazil, France, Germany, or Italy?". Left: Probabilistic Coherence, evaluated using Dutch-Book scenarios in the same chat and across conversations to find violations of probability axioms. Middle: Posterior Fidelity, as KL divergence between LLM posterior and Bayesian oracle. Right: Downstream Task Performance, assessing how non-Bayesian updates affect success in Experimental Design.}
    \label{fig:figure1}
\vspace{-0.53cm}
\end{figure}

To answer this question, we evaluate LLMs along three progressive axes: probabilistic coherence of their beliefs, tested via Dutch-Book constructions \citep{og_DB_introduction} that expose violations of probability axioms; posterior fidelity, measured as divergence of model-generated distributions from a Bayesian oracle; and downstream task performance in adaptive experimental design, to determine whether non-Bayesian drift has practical consequences. We use two complementary environments, \texttt{20 Questions} for internal consistency and our new \texttt{ProteinWorld} framework for recovery of externally determined posteriors, and vary model release year, reasoning, and code execution to isolate which factors drive Bayesian capabilities.

Our main contributions can be summed up as follows:
\vspace{-0.22cm}
\begin{itemize}
    \item A principled evaluation of the extent to which LLMs can approximate Bayesian inference in context, covering internal consistency and recovery of externally determined posteriors.
    \item \texttt{ProteinWorld}, a highly configurable environment that mimics scientific discovery and explicitly benchmarks iterative Bayesian inference.
    \item Evidence that LLMs violate Bayes' rule without reasoning, but that modern reasoning models compute externally determined posteriors exactly, remain consistent over 10+ data points, and, with tools, correctly approximate intractable posteriors, potentially rendering current belief-updating harnesses unnecessary and enabling more capable AI scientists.
\end{itemize}

\vspace{-0.3cm}
\section{Related Work}
\vspace{-0.2cm}
\paragraph{Bayesian inference in LLMs}
There exist some previous approaches looking at in-context Bayesian inference of LLMs. Purpose-trained Transformers can indeed approximate Bayesian inference in context \citep{Transformers_BayInf_1, Transformers_BayInf_2, Transformers_BayInf_3, Transformers_BayInf_4}. For general-purpose LLMs, ICL has been cast as implicit Bayesian inference over latent task concepts \citep{ICL_Is_BayInf}. However, they have been found to violate exchangeability \citep{ICL_nonBayesian_1}, weight evidence by its position in context \citep{ICL_nonBayesian_3}, and being Bayesian only in expectation over context orderings \citep{ICL_nonBayesian_2}. We continue in this line of work by evaluating explicitly stated belief distributions of general purpose LLMs and asking if capabilities like reasoning or tools can serve to close the non-Bayesian gap.
 \vspace{-0.2cm}
\paragraph{Probabilistic coherence and calibration}
Before looking at the quality of LLM-based belief updates, it is important to clarify whether LLMs can work with probabilities. \citet{LLM_DutchBook_Theory} argues that LLM-sampled linguistic probabilities preclude ideal rationality, and \citet{LLM_DutchBook_Practice} find incoherent judgments empirically, though only in two-event toy settings, with further coherence audits emerging recently \citep{LLM_DB_1, LLM_DB_2, LLM_DB_3, LLM_DB_4}. Calibration work instead scores stated confidences against outcomes \citep{lin2022teaching, kadavath2022language, verbalize_probabilities, xiong2024canllms, halawi2024forecasting, forecastbench}, an external and orthogonal criterion. We test coherence only so that violations reflect failures of internal logic rather than world knowledge, and contrast cross- with single-dialogue elicitation, a dimension largely absent from prior work.
 
\paragraph{Measuring belief updates against a Bayesian oracle}
Some previous work looks at the quality of LLM-based belief updates. Existing evaluations differ in where the reference posterior originates. With a fully specified generative process, the reference is exact: biased-coin sequences \citep{ProtWorld_comp_1}, verbal probabilistic reasoning problems \citep{ProtWorld_comp_2, ProtWorld_comp_3, ProtWorld_comp_4}, and, concurrent with our work, multi-turn environments with closed-form references \citep{bayesbench}. Our work introduces \texttt{ProteinWorld}, which adds out-of-distribution procedural worlds, engineered analytical intractability over continuous latent attributes, and a downstream metric to analyze cost of deviations. Absent a ground-truth model, the reference must be built from the model's own quantities: \citet{20Q_comparison_noKL} score \texttt{20Q} updates via proxies such as falsely retained hypotheses, while \citet{20Q_comparison_KL} measure KL divergence against products of self-elicited priors and likelihoods over four fixed hypotheses. We expand on this by analyzing the self-consistency paradigm with an open, dynamically evolving hypothesis and experiment space.
\vspace{-0.2cm}

\paragraph{Improving belief updates}
Due to the assumption of low-quality belief updates, previous work has either utilized strict harnesses \citep{ca_bed} or finetuning \citep{Deepmind_BayesianAgent} to force Bayesian updates in sequential tasks. The harnessed approach is exemplified by \citet{bedllm}, the work most closely related to ours, since we build directly on their \texttt{20Q} protocol. We differ from this line of work by evaluating the premise itself, and treating access to reasoning and code execution as the intervention.

\vspace{-0.2cm}
\section{Background}\label{sec:background}
\vspace{-0.2cm}
\paragraph{Bayesian inference}
A key capability of a Bayesian agent is Bayesian inference: the process of updating beliefs over a set of hypotheses $\Theta$ in light of new evidence. In scientific discovery, evidence arrives sequentially \citep{modernBED}: starting from a prior $p_0$, the agent performs experiments $\xi_i$, observes outcomes $y_i$, and accumulates a history $h_{1:k}$ with $h_i=(\xi_i, y_i)$. Given a likelihood model $p(y \mid \theta, \xi)$ specifying how probable outcome $y$ is under hypothesis $\theta$ and experiment $\xi$, Bayes' rule \citep{bayes1764essay} yields the posterior after the $k$-th experiment: The posterior after experiment $k-1$ serves as the prior for experiment $k$, and assuming outcomes are conditionally independent given $\theta$ and the experiment, the update takes the form
\vspace{-0.2cm}
\begin{equation}
\label{eq:sequential_bayes_update}
p_{k}(\theta) := p(\theta \mid h_{1:k}) \propto p(y_{k} \mid \theta, \xi_{k})\, p_{k-1}(\theta) \propto p_0(\theta) \prod_{i=1}^{k} p(y_{i} \mid \theta, \xi_i).
\vspace{-0.2cm}
\end{equation}
While this is a closed-form equation in principle, finding a solution is often analytically intractable in practical settings, due to the complex structure of $p(y \mid \theta, \xi)$ or $\Theta$ in the normalizing integral. It therefore remains a challenging problem, making an analysis of current LLM capabilities in solving it interesting since it goes beyond basic mathematical calculations.

We analyze the posterior distribution over $\Theta$ generated by the LLM either through verbalization \citep{verbalize_probabilities} or by using token-logprobs when being passed prior and experimental data in context. We denote these LLM-elicited probabilities as $p_\text{LLM}^{k}:= p_{\text{LLM}}(\theta \mid h_{1:k})$, and compare the distribution to the rational Bayesian posterior $p_\text{oracle}^{k}$ calculated following Eq.~\ref{eq:sequential_bayes_update}.

\vspace{-0.2cm}
\paragraph{Dutch Books}
If an agent's beliefs over $\Theta$ are inconsistent with core probability axioms, they can be exploited via a Dutch Book. A Dutch Book is a collection of simultaneous bets formed from an agent's stated subjective probabilities that yields a guaranteed loss, regardless of the realized outcome \citep{og_DB_introduction}. This relies on the fair-price assumption: if the agent assigns probability $q_j = p(H_j)$ to event $H_j$, then they should be willing to buy a unit contract paying $1$ if $H_j$ occurs and $0$ otherwise for at most $q_j$, or sell it for at least $q_j$ \citep{newer_DB_introduction}.

Beyond the formal definition in App.~\ref{app:DB_payoff_formula}, consider the simple example $\Theta = \{A,B\}$ with disjoint events $H_1=A$ and $H_2=B$, and suppose the agent assigns $q_1=0.6$ and $q_2=0.5$. Selling the agent a unit contract on both those events yields a guaranteed net loss, since it pays $1.1$ up front, while receiving exactly $1$ in the end. In this case, the probabilities sum to $1.1$, violating Kolmogorov's normalization axiom \citep{kolmogorov}. More generally, the existence of Dutch Books in declared beliefs is directly equivalent to the violation of one of Kolmogorov's core axioms \citep{sep-dutch-book, reverse_DB, reverse_DB_2}, making them an appropriate tool to check whether stated beliefs can even be interpreted as a probability distribution.

\vspace{-0.1cm}
\paragraph{Experimental Design}
Experimental design addresses the question of \textit{which} experiment $\xi$ out of a set of available experiments $\Xi$ an agent should perform next to learn as much as possible about its hypotheses. The Bayesian approach frames the choice of experiment as a decision problem: given current beliefs $p(\theta)$ over $\Theta$, the agent should select the experiment $\xi \in \Xi$ that maximizes the expected utility of its outcome \citep{lindley1956measure, chaloner1995bayesian}. Following \citet{lindley1956measure}, the canonical utility is the information gained about $\theta$, leading to the Expected Information Gain (EIG), which can equivalently be written as the expected reduction in entropy over the hypotheses or over the experimental outcomes:
\begin{equation}
\operatorname{EIG}(\xi) \;=\; \mathbb{E}_{y \sim p(y \mid \xi)} \!\left[ \mathrm{H}\!\left[p(\theta)\right] - \mathrm{H}\!\left[p(\theta \mid \xi, y)\right] \right] \;=\; \mathbb{E}_{\theta \sim p(\theta)}\!\left[\mathrm{H}\!\left[p(y \mid \xi)\right] - \mathrm{H}\!\left[p(y \mid \xi, \theta)\right] \right]
\end{equation}
In sequential or adaptive experimental design, this selection step is interleaved with belief updating: Given a hidden ground truth $\theta^\star$, the agent chooses the maximally informative experiment $\xi_k$ based on its current beliefs $p_{k-1}(\theta)$, observes the outcome $y_k \sim p(\cdot \mid \theta^\star, \xi_k)$, updates its beliefs to $p_{k}(\theta \mid h_{1:k})$, and repeats \citep{ryan2016review, modernBED}. This iterative loop of design, observation, and inference formalizes the scientific discovery process, and its efficiency depends critically on the quality of belief updates made by the agent: since the EIG is computed under the agent's current beliefs, belief updates propagate directly into future experiment choices, potentially degrading performance long-term. Since Bayesian updating is optimal here \citep{wald1947complete, zellner1988optimal}, running this loop with an LLM agent against a Bayesian oracle lets us quantify belief divergence and resulting cost in discovery performance.

\section{Methodology}\label{sec:methodology}
The goal of our work is to answer the central research question posed in Sec.~\ref{sec:introduction}: \textit{To what extent can LLMs perform or approximate Bayesian inference in context?} Answering this question requires more than a single accuracy metric, since in-context belief updates can fail in qualitatively different ways. We therefore evaluate in-context Bayesian inference along three progressive axes that span from foundational probability theory to practical application.
\vspace{-0.2cm}

\subsection{Probabilistic Coherence}\vspace{-0.1cm}
A necessary prerequisite to interpreting the outputs of an LLM as posterior distributions is that they are \textit{coherent} with the basic laws of probability theory. If a model assigns mutually inconsistent probabilities, or fails to define a valid distribution over hypotheses, any direct comparison to a Bayesian oracle becomes conceptually limited. As discussed in Sec.~\ref{sec:background}, the existence of Dutch Books is directly equivalent to beliefs being incoherent. We use binary judgmental forecast questions from ForecastBench \citep{forecastbench} (\href{https://github.com/forecastingresearch/forecastbench-datasets/blob/main/datasets/question_sets/2026-06-21-llm.json}{June 21, 2026 version}) about future political, economic, technological, and social events as basis to elicit probabilities. Each binary ForecastBench question is interpreted as an event $A \subseteq \Theta$ within a broader outcome space $\Theta$. For example, the question ``Will Spider-Man: Beyond the Spider-Verse get a $95$\% or higher on Rotten Tomatoes?'' defines the event $A=\{\theta^\star \geq 95\}$, where $\theta^\star$ denotes the eventual Rotten Tomatoes Rating. We elicit probabilities for collections of related events, checking for violations of extensionality, normalization, and monotonicity. The precise Dutch Book constructions for each violation type is in App.~\ref{app:DB_three_scenarios}.

We consider two elicitation settings. First, we query each event independently in a separate chat. This setting tests whether the model's responses across independent interactions can be represented by a single coherent probability distribution. Second, we elicit the same set of probabilities jointly within a single chat, checking the weaker hypothesis of the model maintaining probabilistic coherence within one context. Fig.~\ref{fig:figure1} (left) illustrates one example of this procedure.
\vspace{-0.2cm}

\subsection{Sequential belief updating}\vspace{-0.1cm} Following probabilistic coherence, we move to the scientific discovery loop from Sec.~\ref{sec:background}. Since the \textit{rational} posterior is only well-defined relative to a prior $p_0$, we fix that externally. In practice, $p_0$ could be derived from existing literature; for evaluating belief updates, it is only important that all agents share the same $p_0$. We run three trajectories: We use the LLM to update beliefs ${p}^{\,k}_\text{LLM}$ and choose experiments. Then, we compute a passive oracle trajectory based on this run: $p^{\,k}_\text{ora-pas}$ is obtained by applying Eq.~\ref{eq:sequential_bayes_update} with $p_0$ and the experimental results of the first runs. A rational agent ought to hold this belief state given the model's evidence. Third, we let an active Bayesian agent update beliefs $p^{\,k}_\text{ora-act}$ using Eq.~\ref{eq:sequential_bayes_update} and choose experiments based on those beliefs.
\vspace{-0.3cm}

\paragraph{Posterior fidelity}
Given the trajectories $p^{k}_\text{LLM}$ and $p^{\,k}_\text{ora-pas}$, we analyze whether the LLM moves its beliefs as the evidence warrants, following the principles of Bayesian inference. We compare the two distributions directly, using the KL divergence \citep{KL_divergence}:
\begin{equation}
    \text{KL}(p^{\,k}_\text{ora-pas} \parallel p_{\text{LLM}}^{k}) = \int_{\theta \in \Theta}p^{\,k}_\text{ora-pas}(\theta) \cdot \log\frac{p^{\,k}_\text{ora-pas}(\theta)}{p_{\text{LLM}}^{k}(\theta)}d\theta
\end{equation}
Across iterations of the scientific discovery workflow, this traces the trajectory of non-Bayesian divergence as evidence accumulates. We deliberately use the forward KL to discourage early discarding of hypotheses that are still valid according to the oracle, as that is arguably a bigger problem in scientific discovery than not immediately discarding invalid hypotheses (measured by the reverse KL \citep{KL_forwardvsbackward}). Fig.~\ref{fig:figure1} (middle) illustrates the procedure for one iteration with three executed experiments. To prevent division by $0$, we clip $p^{k}_\text{LLM}$ from below at $10^{-6}$ and re-normalize.
\vspace{-0.3cm}

\paragraph{Downstream task performance}
Beyond deviations from normative probability theory, a critical question remains: does non-Bayesian belief updating inherently degrade an agent's functional utility? In scientific discovery, the central issue is not whether $p_{\text{LLM}}^{k}$ violates principles of Bayesian probability theory, but whether relying on the model's beliefs leads to worse experimental decisions and performance. The comparison between $p^k_\text{LLM}$ and $p^{\,k}_\text{ora-act}$ captures exactly this. We score each trajectory by the mass it places on the hypothesis the environment actually drew, $\{p_k(\theta^\star)\}_{1:T}$, the quantity a discovery procedure exists to raise. We also report the posterior-mass gap
\begin{equation}\label{eq:mass_gap}
  p^{\,k}_\text{ora-act}(\theta^\star) - p^k_\text{LLM}(\theta^\star) ,
\end{equation}
since it is less influenced by how hard the environment is than absolute identification accuracy. Because its two terms come from independently generated histories, as visualized in Fig.~\ref{fig:figure1} (right), this difference is the desired end-to-end performance difference. On a single run, a model that collapses its mass onto one hypothesis after a single experiment may happen to be right and beat the oracle. However, over the long term, this kind of behavior will also lead to sometimes falsely eliminating $\theta^\star$, leading to the oracle outperforming on average \citep{bayesian_decisiontheoretic, modernBED}.
\vspace{-0.4cm}

\section{Experimental Setup}
To answer our central research question, we need to concretely instantiate the evaluation framework introduced in Sec.~\ref{sec:methodology}. We propose two sequential belief updating environments deliberately designed to be complementary in their analysis, and present the utilized configurations for our evaluation.
\vspace{-0.2cm}

\subsection{Environments for sequential belief updating}\vspace{-0.1cm}\label{sec:environments}
First, we use \texttt{20Q}, building on the approach of \citet{bedllm}. An experiment is a single Yes/No question, e.g. $\xi = \text{"Does it swim?"}$, $y \in \{\text{Yes}, \text{No}\}$. We make no restriction on the type of Yes/No questions that can be asked: $\Xi = \{\text{All Yes/No questions}\}$. There is also no restriction on animals that can be guessed: $\Theta = \{\text{All existing animals\}}$, since the unrestricted experiment space would otherwise make the game trivially solvable by filtering half of $\Theta$ at every iteration. In this scenario, both $\Theta$ and $\Xi$ are arbitrarily large and unstructured, but the likelihood model $p(y \mid \theta, \xi)$ is very simple, with $y$ always being a discrete random variable with the same two possible outcomes.

Second, we use our new synthetic \texttt{ProteinWorld} scenario. It was designed to complement \texttt{20Q} and mimic the scientific discovery process of protein classification, using measurements of latent attributes to classify a subject. To ensure the tasks are strictly out-of-distribution for the LLM, we procedurally generate a completely new "world" for each instance. This involves using an LLM to create fictional terminology for protein classes, attributes, and experiments, and assigning randomized probability distributions to govern their relationships. \texttt{ProteinWorld} is highly configurable: evaluators have full control over the number of classes, attribute dependencies, and the complexity of the observation models, allowing the benchmark to represent a wide spectrum of difficulty. \\
Formally, the hypothesis space $\Theta = \{\theta_1, \dots, \theta_M\}$ is a finite, explicitly defined set of distinct protein classes. Each class $\theta \in \Theta$ is characterized by a latent attribute vector $\mathbf{a} \in \mathcal{A}$ governed by a class-conditional distribution $p(\mathbf{a} \mid \theta)$ featuring randomized parameters, where different attributes might depend on each other. The action space $\Xi = \{\xi_1, \dots, \xi_N\}$ consists of a fixed, finite set of experiments. Executing an experiment $\xi \in \Xi$ yields an observation $y \in \mathcal{Y}_\xi$ according to an observation model $p(y \mid \mathbf{a}, \xi)$. \texttt{ProteinWorld} therefore pairs simple, discrete sets $\Theta$ and $\Xi$ with highly complex observation dynamics, where the likelihood model \begin{equation}
\label{eq:likelihood_protworld}
p(y \mid \theta, \xi) = \int_{\mathcal{A}} p(y \mid \mathbf{a}, \xi) p(\mathbf{a} \mid \theta) \, d\mathbf{a},
\end{equation}
can be engineered to be as complex as desired, since both $ p(y \mid \mathbf{a}, \xi)$ and $p(\mathbf{a} \mid \theta)$ are fully in the hand of the evaluator. We use a \texttt{simple} version, where the likelihood admits a closed-form analytical solution, and a \texttt{hard} version, where the likelihood has to be approximated using probabilistic methods like MC simulation. Instances of these two difficulty classes are provided in App.~\ref{app:ProteinWorld_example_instances}.
\vspace{-0.3cm}

\paragraph{LLM posterior}
For \texttt{20Q}, the hypothesis space $\Theta$ is infinite, meaning it is intractable for an LLM to generate an explicit probability distribution over all $\theta \in \Theta$. We therefore maintain a dynamic subset $\Theta_k \subset \Theta$, such that $p_\text{LLM}^k(\theta) \neq 0 \Leftrightarrow \theta \in \Theta_k$. This subset is filtered and extended once per iteration, a necessary deviation from strict Bayesian updating to handle the infinite hypothesis space. We then model $\Theta=\Theta_k$. To compute $p_\text{LLM}^k$, we make the LLM evaluate the compatibility of each hypothesis $\theta \in \Theta_k$ with the experimental history $h_{1:k}$. The extracted probabilities are then normalized across $\Theta_k$ to form a valid distribution. Note that this procedure naturally induces a Uniform prior over $\Theta_k$, since the compatibility of all hypotheses with an empty history is perfect.

For \texttt{ProteinWorld}, $\Theta$ is fixed and finite, so the generation of the LLM probability distribution is much simpler: At iteration $k$, the LLM is given the world setup including the exact distributions and parameters for $p(\mathbf{a} \mid \theta)$ and $p(y \mid \mathbf{a}, \xi)$, as well as all observed experiments and results $(h_1,\dots,h_k)$, in context. It is also given an explicit prior $p_0$ assigning probabilities to all $\theta \in \Theta$. The LLM is then asked to output the posterior probability distribution across $\Theta$.
\vspace{-0.3cm}

\paragraph{Oracle posterior} 
For \texttt{20Q}, the prior is, as discussed, a Uniform distribution over $\Theta_k$. Following Eq.~\ref{eq:sequential_bayes_update}, we also need the experimental likelihoods to calculate $p^k_\text{oracle}$. The open-ended structure of $\Theta$ and $\Xi$ prohibits fully specifying $p(y \mid \theta, \xi)$ using an external ground truth. Instead, we define experimental likelihoods via the LLM's own assessments. This evaluates the LLM's \textit{internal probabilistic consistency}: Specific probabilities assigned by the LLM to question-answer pairs do not matter, only whether the model's aggregated posterior after a sequence of events aligns with exact Bayesian updates derived from its own marginal likelihoods. 

In contrast, \texttt{ProteinWorld} evaluates the LLM's ability to recover an \textit{externally determined} posterior: All of prior $p_0$, attribute distribution $p(\mathbf{a} \mid \theta)$, and the observation model $p(y \mid \mathbf{a}, \xi)$ are mathematically specified, meaning the experimental likelihood  $p(y_k \mid \theta, \xi_k)$ can be exactly calculated following Eq.~\ref{eq:likelihood_protworld}. This is done analytically for \texttt{simple} and using a MC estimator for \texttt{hard}.
\vspace{-0.3cm}

\subsection{Models and Evaluation Protocol}\label{sec:exp_setup}\vspace{-0.1cm}
\paragraph{Probabilistic Coherence}
For each of the three violation types, we evaluate 10 different ForecastBench scenarios. We elicit Dutch Books on \texttt{Gemini-2.5-Pro} \citep{gemini-2.5}, \texttt{Gemma-4-31B-it} \citep{gemma-4}, and \texttt{Qwen-2.5-32B-Instruct} \citep{qwen2.5} in order to cover both smaller open-source and larger proprietary language models, a range of release years, and both reasoning models as well as a non-reasoning model.
\vspace{-0.3cm}

\paragraph{Sequential Belief Updating}
We evaluate \texttt{20Q} on $40$ target animals $\theta^\star$ (see App.~\ref{app:20Q_animallist}), using $T=20$. For \texttt{ProteinWorld}, we generate $5$ independent worlds; each world contains $|\Theta|=8$ protein classes, characterized by attribute vectors of dimension $8$, together with $|\Xi|=6$ available experiments. Within each world, every protein class serves as the ground-truth hypothesis $\theta^\star$ exactly once. This yields $40$ runs in total, using $T=6$. We analyze sequential belief updating on \texttt{Qwen-2.5-32B-Instruct}, \texttt{Phi-4}~\citep{phi-4}, \texttt{Gemma-4-31B-it}, \texttt{Gemma-4-26B-A4B-it} \citep{gemma-4}, and \texttt{Qwen-3.6-35B-A3B} \citep{qwen36_35b_a3b}, spanning a range of release years, number of parameters and model families. All five models are evaluated once without reasoning, and the latter three models, which support a reasoning mode, are also evaluated with reasoning. Finally, \texttt{ProteinWorld - hard} is also evaluated using \texttt{Gemini-2.5-Pro} with the native Python code-execution tool \citep{gemini_code_execution} available.
\vspace{-0.2cm}

\section{Results \& Discussion}
\vspace{-0.2cm}
We now answer our central research question: \textit{To what extent can LLMs perform or approximate Bayesian inference in context?} We apply the framework of Sec.~\ref{sec:methodology} to the environments of Sec.~\ref{sec:environments} and decompose the question into a sequence of sub-questions, each building on the result of the last.
\vspace{-0.7cm}
\paragraph{Are LLM-elicited beliefs probabilistically coherent?}
\begin{wrapfigure}{r}{0.7\linewidth}
\centering
\includegraphics[width=\linewidth]{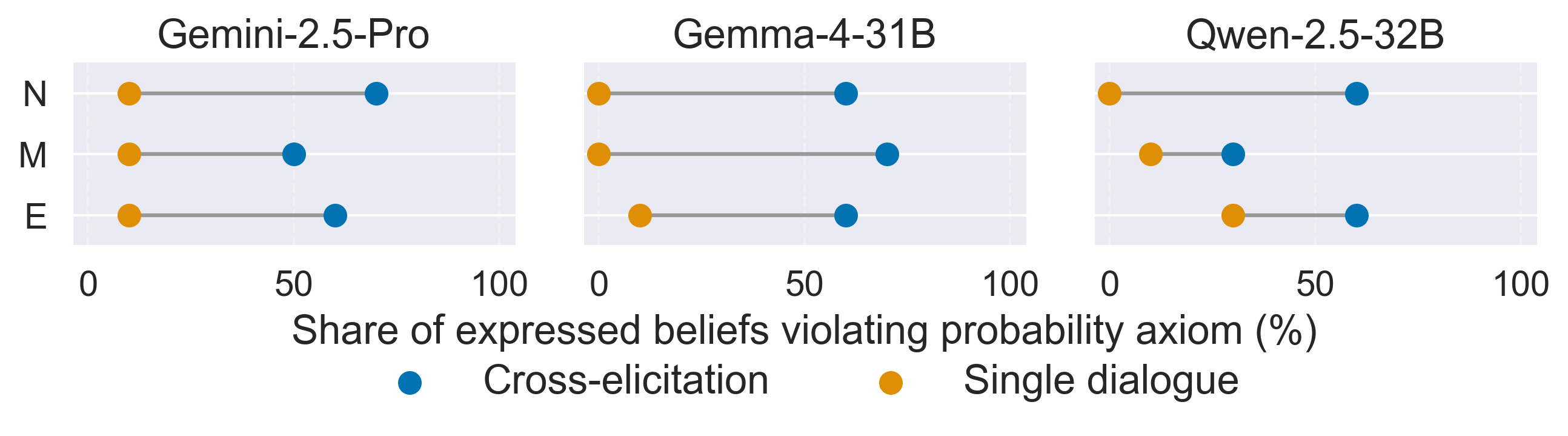}
\caption{\textbf{Share of elicited probability judgments that are Dutch Bookable under cross-conversation and within-dialogue elicitation.} While beliefs elicited across different conversations frequently violate normalization (N), monotonicity (M), and extensionality (E), stronger LLMs are mostly coherent in the within-dialogue setting.}
\label{fig:db_cross_vs_intra_convo}
\end{wrapfigure}
App.~\ref{app:DB_moreexamples} shows Dutch Books constructed from judgments elicited across separate conversations, demonstrating that such judgments need not be jointly representable by a single probability measure. Fig.~\ref{fig:db_cross_vs_intra_convo} shows that this depends strongly on the elicitation setting. Across conversations, violation rates exceed 50\% of the 30 scenarios for almost all model-constraint combinations. Within a single dialogue, they drop markedly for all three models and axioms, to 10\% for Gemini, and to a single scenario out of 30 for Gemma. Access to the preceding judgments therefore appears sufficient to enforce consistency, suggesting that cross-conversation incoherence reflects the absence of a persistent common representation rather than an inability to apply elementary probability rules.

\begin{boxedresult}\label{res:probab_axioms_good}
When probabilities are elicited within a single dialogue, modern reasoning LLMs generally satisfy basic probability axioms, and their outputs can be treated as evaluations from a coherent probability distribution within the scope of these tests.
\end{boxedresult}

While these tests do not establish universal within-dialogue coherence, they suffice for what follows: In our scientific-discovery setting, the model receives hypotheses, priors, and evidence in one dialogue and returns a posterior in that same context. The strong within-dialogue performance observed therefore supports treating the outputs as valid probability assignments.
\vspace{-0.2cm}

\paragraph{Can LLMs perform good approximate Bayesian inference in context?}
\begin{figure}[b]
\centering
\includegraphics[width=0.9\linewidth]{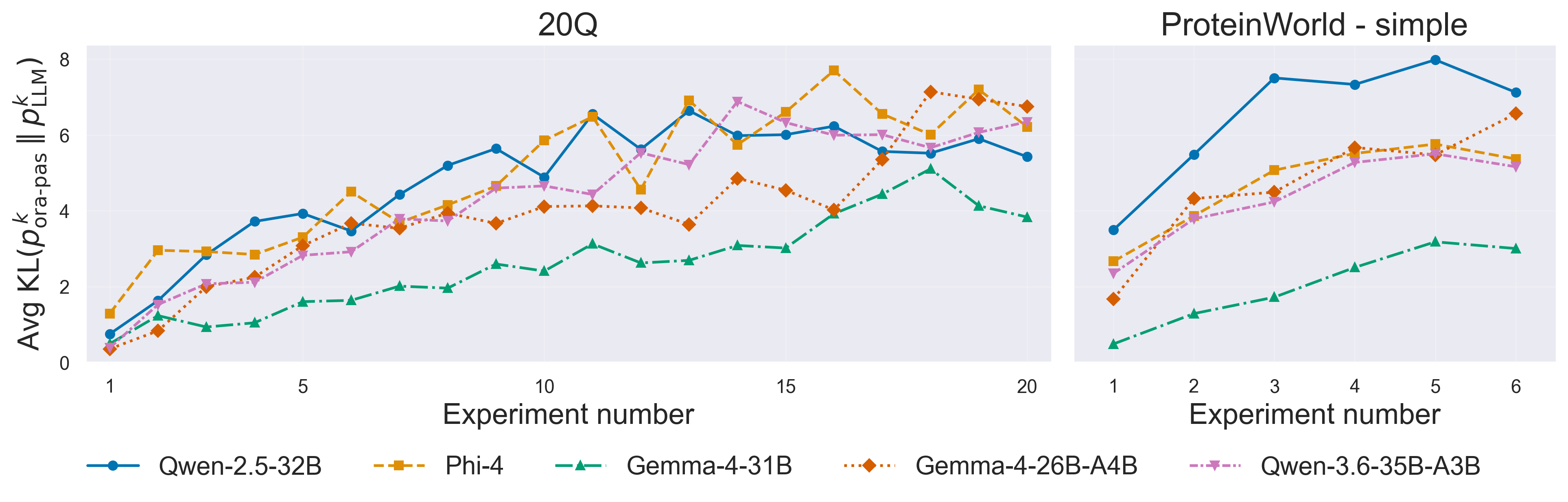}
\caption{\textbf{Posterior fidelity of non-reasoning LLMs on \texttt{20Q} and \texttt{ProteinWorld - simple}.} On both environments, the average $\text{KL}({p}_{\text{ora-pas}}^{\,k} \parallel p_{\text{LLM}}^{k})$ by number of experiments $k$ takes on large values after only a few experiments have been executed, showing that LLM belief updates diverge substantially from those presented by the Bayesian oracle with same evidence.}
\label{fig:no_reasoning_posterior_distribution_acc}
\end{figure}
Having shown that LLMs can be used to construct valid posteriors, we now ask how good those distributions are in the context of Bayesian inference. Fig.~\ref{fig:no_reasoning_posterior_distribution_acc} reports our evaluation of posterior fidelity on \texttt{20Q} and \texttt{ProteinWorld} for a range of LLMs \textit{without reasoning}. Across model families and release years, the results show that LLMs do not perform accurate Bayesian updates in either environment. They neither maintain internal probabilistic consistency over a substantial number of experiments nor accurately compute the closed-form posterior in the \texttt{ProteinWorld - simple} setting.

\begin{figure}[t]
\centering
\includegraphics[width=0.9\linewidth]{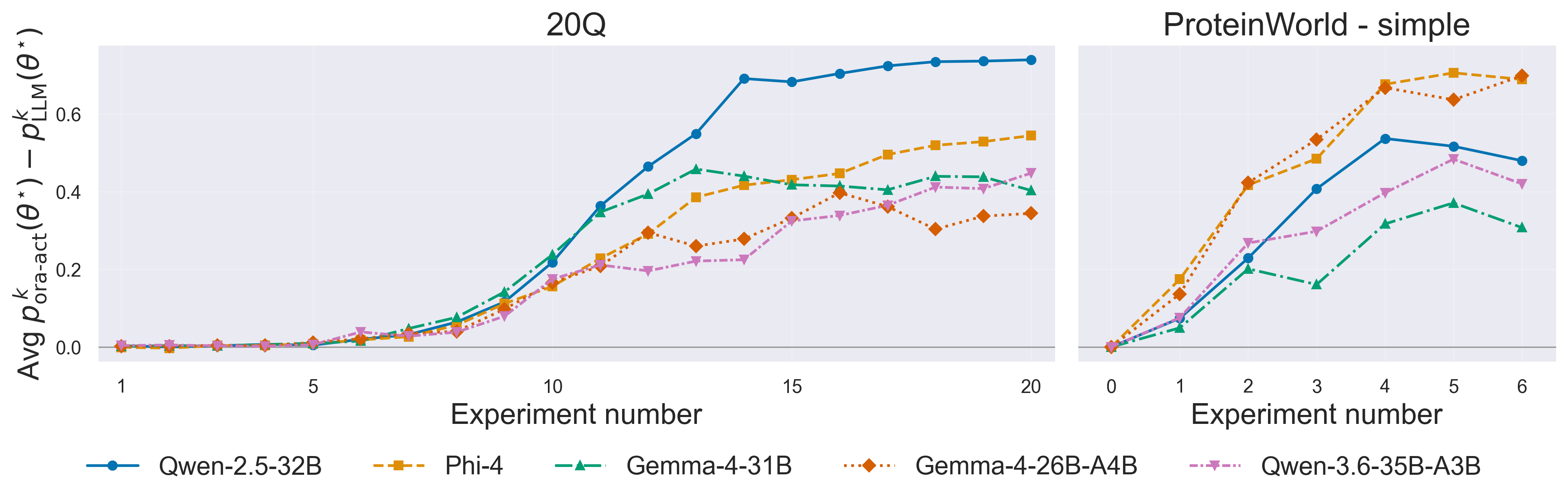}
\vspace{-0.4cm}
\caption{\textbf{Downstream task performance of different non-reasoning LLMs on \texttt{20Q} and \texttt{ProteinWorld - simple}}. On both environments, the average ${p}_{\text{ora-act}}^{\,k}(\theta^\star) - p_{\text{LLM}}^{k}(\theta^\star)$ by number of experiments $k$ rises to above $30$\% for all LLMs, showing that the non-Bayesian belief updates lead to substantially decreased performance in experimental design.}
\label{fig:no_reasoning_downstream}
\vspace{-0.5cm}
\end{figure}

This deviation from correct Bayesian updates is shown in Fig.~\ref{fig:no_reasoning_downstream} to have major consequences in the downstream task performance: LLM-based probability updates reduce final success rates by at least $30$\% across all models and both environments when compared to the active oracle trajectory. On \texttt{20Q}, these results are even worse for the weaker models, with catastrophic collapses to close to $0$\% final accuracy. Qualitatively, on \texttt{ProteinWorld}, modern models with no access to reasoning frequently identify the most likely protein correctly, but assign too much or too little mass to it, while weaker models show no signal beyond random guessing (App.~\ref{app:qual_simple_noreasoning}).

\begin{boxedresult}\label{res:BayesInf_incontext_bad}
Without reasoning, LLMs fail to perform accurate Bayesian inference in context, neither maintaining internal probabilistic consistency nor computing simple analytical posteriors reliably, which markedly impacts downstream task performance.
\end{boxedresult}
\vspace{-0.3cm}

\paragraph{How much does reasoning improve in context Bayesian inference?}
\begin{figure}[b]
\centering
\begin{subfigure}[t]{0.49\linewidth}
    \centering
    \includegraphics[width=\linewidth]{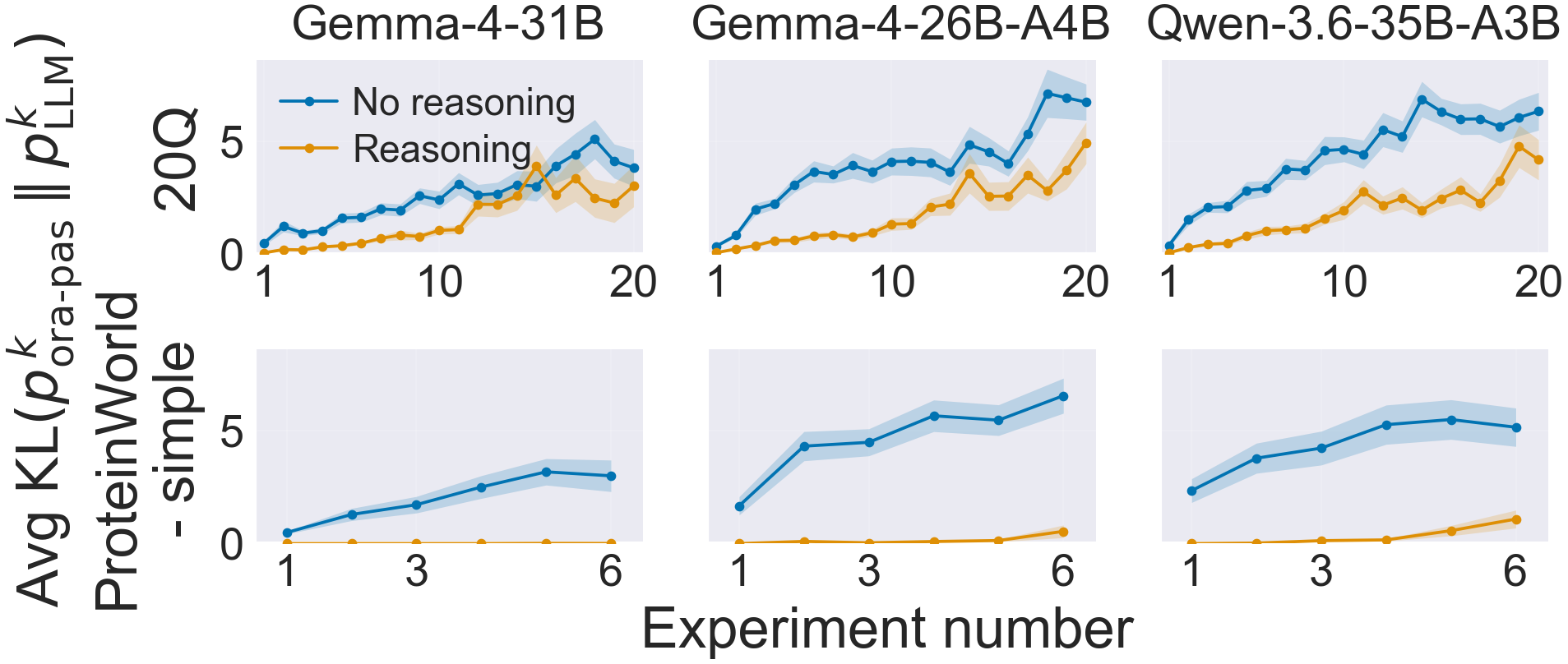}
    \label{fig:reasoning_posterior_acc}
\end{subfigure}
\hfill
\begin{subfigure}[t]{0.49\linewidth}
    \centering
    \includegraphics[width=\linewidth]{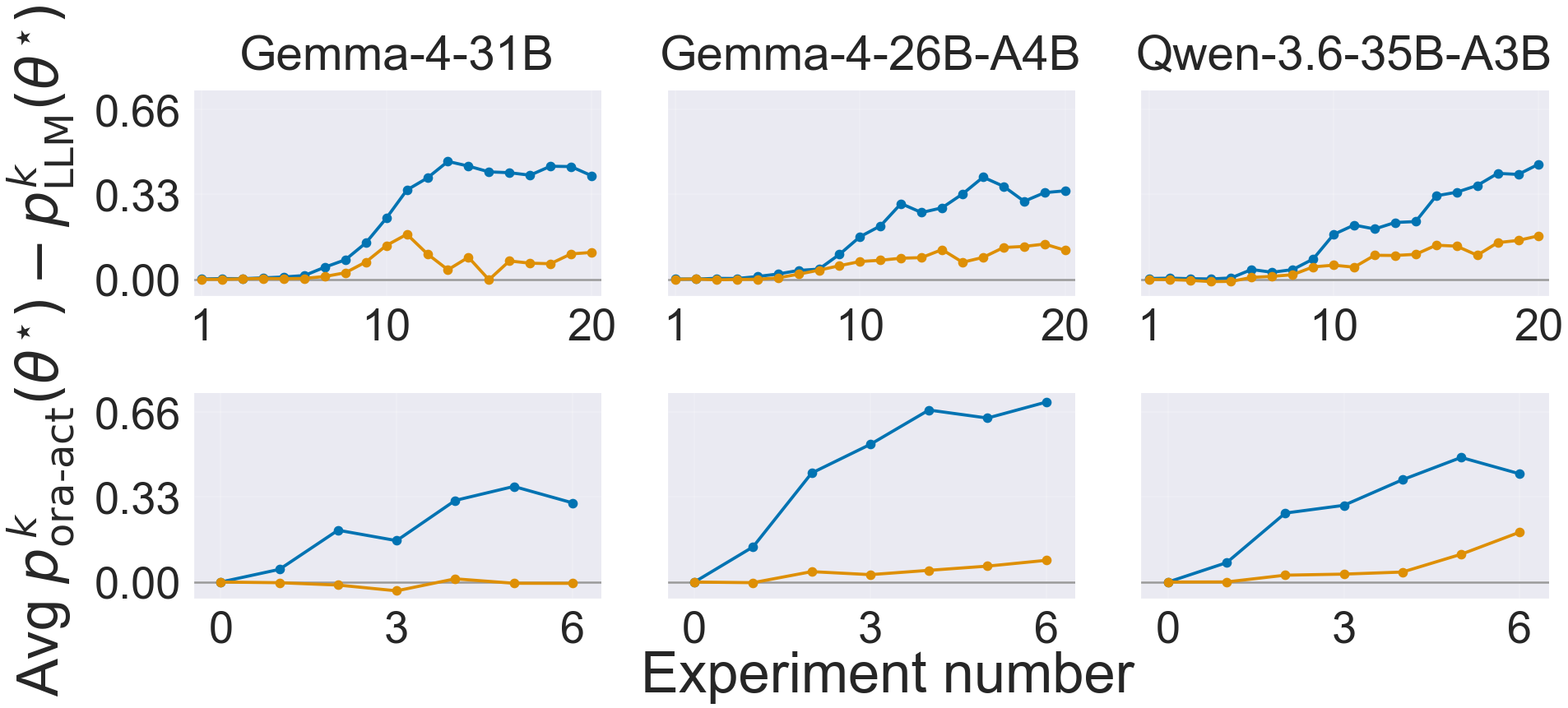}
    \label{fig:reasoning_downstream_perf}
\end{subfigure}
\vspace{-0.2cm}
\caption{\textbf{Impact of native reasoning on in-context Bayesian inference for modern open-source LLMs on \texttt{20Q} and \texttt{ProteinWorld - simple}}. We display, by number of executed experiments $k$, with reasoning enabled and disabled: Left: Posterior fidelity - average $\text{KL}(p^{\,k}_\text{ora-pas} \parallel p_{\text{LLM}}^{k})$, Right: Downstream task performance - average posterior-mass gap ${p}^{\,k}_\text{ora-act}(\theta^\star)-p^k_\text{LLM}(\theta^\star)$. Reasoning models show greatly reduced losses compared to their non-reasoning counterparts, providing evidence that the flexible compute budget introduced through reasoning is effective in improving belief update quality and even leads to rivaling oracle performance.}
\label{fig:reasoning_comparison}
\vspace{-0.5cm}
\end{figure}
For Res.~\ref{res:BayesInf_incontext_bad}, all probabilities are obtained using a fixed compute budget of exactly the tokens needed to output probabilities. Taking a step further, we now allow models to dynamically choose a higher computation budget for themselves by activating their native reasoning modes, and analyze the impact on Bayesian updating capabilities. The results are shown in Fig.~\ref{fig:reasoning_comparison}.

On \texttt{20Q}, reasoning yields a substantial improvement for all models. This shows in the posterior fidelity: reasoning strongly reduces the KL divergence across all 3 models, and \texttt{Gemma-4-31B-it} even achieves mostly consistent updates for up to 10 experiments. The downstream task results corroborate the practical value of these better-calibrated updates: the difference between oracle and LLM performance is substantially lower for all models when enabling their reasoning abilities. 

This observation holds even stronger for \texttt{ProteinWorld - simple}. Modern reasoning models achieve a KL close to $0$ regardless of the number of experiments. This is reflected in the downstream task performance, where \texttt{Gemma-4-31B-it} manages to match the performance of the oracle across the entire discovery process. Qualitatively, the models reconstruct likelihoods and Bayes' rule correctly; errors stem from numerical slips rather than conceptual failures. The correct posterior is typically re-verified for thousands of redundant tokens (App.~\ref{app:qual_analysis_simple_reasoning}).

\begin{boxedresult}\label{res:BayesInf_incontext_reasoning_good}
Reasoning substantially improves the ability of modern LLMs to perform Bayesian inference in context, enabling computation of simple externally determined posteriors and internal probabilistic consistency over experiment sequences of up to 10 datapoints. This allows for good downstream task performance closer to the optimum presented by Bayesian oracles.
\end{boxedresult}

\begin{wrapfigure}{r}{0.55\linewidth}
\centering
\includegraphics[width=\linewidth,keepaspectratio]{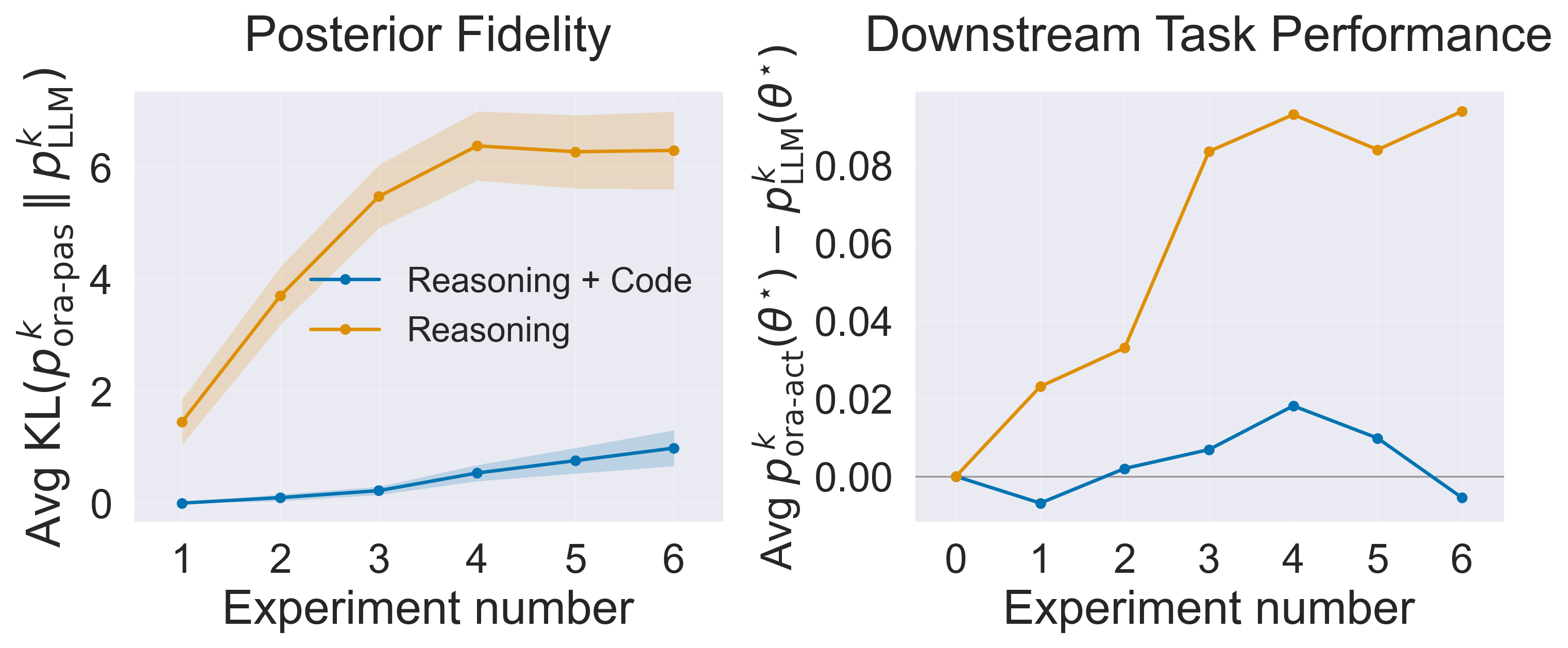}
\caption{\textbf{Impact of code execution on in-context Bayesian inference for \texttt{Gemini-2.5-Pro} on \texttt{ProteinWorld - hard}.} We display, by number of executed experiments $k$, a baseline with reasoning only against reasoning + code execution: Left: Posterior fidelity, Right: Downstream task performance. Access to a Python environment drives KL divergence and performance gap close to $0$, showing that the model can use tools to approximate an analytically intractable posterior effectively.}
\label{fig:gemini_Python_good}
\end{wrapfigure}

These positive results do not carry over to \texttt{ProteinWorld - hard}, where likelihoods do not admit a closed-form solution: While models still set up the Bayesian model correctly and recognize unprompted that the marginalization in Eq.~\ref{eq:likelihood_protworld} is analytically intractable, they proceed as though they could run a Monte Carlo procedure and report hallucinated simulation output, or make too broad numerical approximations (App.~\ref{app:qual_analysis_hard_reasoning}). Fig.~\ref{fig:gemini_Python_good} shows the resulting high KL divergence and performance degradation compared to the oracle for \texttt{Gemini-2.5-Pro}.
\vspace{-0.2cm}

\paragraph{Can LLMs adapt to intractable posteriors with additional tools?} 
Having found that LLMs with reasoning correctly classify numerical simulation as appropriate for \texttt{ProteinWorld - hard}, we now check whether they will act accordingly when given access to tools enabling such simulation. We therefore allow \texttt{Gemini-2.5-Pro} use its native Python environment.

Fig.~\ref{fig:gemini_Python_good} shows the model can exploit code execution to perform good approximate Bayesian inference in the \texttt{ProteinWorld - hard} setting. This is reflected both in the posterior fidelity and in the downstream task performance, where it matches the oracle. Qualitatively, the model performs the same analysis as before and correctly acts on its findings by writing a numerically stable Monte Carlo estimator (App.~\ref{app:qual_analysis_hard_code}).

\begin{boxedresult}\label{res:Bayes_incontext_reasoning_and_tools}
With access to external tools such as code execution environments, modern LLMs can select appropriate tools for Bayesian inference tasks and use them to approximate even analytically intractable posteriors effectively in context, enabling strong results in both posterior fidelity and downstream task performance.
\end{boxedresult}
\vspace{-0.3cm}

\paragraph{Are current strict harnesses for LLM belief updating necessary?}
Driven by our findings, we re-evaluate current belief-updating harnesses: state-of-the-art frameworks like BED-LLM \citep{bedllm} assume bad probability judgments by LLMs, and fix large parts of the probabilistic model externally. We check whether these strong restrictions are still necessary in App.~\ref{app:bed_llm_analysis}.

\begin{boxedresult}\label{res:harnesses_unnecessary}
For modern reasoning LLMs, the structural harnesses employed by recent belief-updating frameworks such as BED-LLM may no longer be necessary, with full belief updates by the LLM itself reaching comparable downstream task performances.
\end{boxedresult}
\vspace{-0.2cm}

Given our findings, we believe that the correct approach to AI scientists and LLM-based Bayesian inference are less restrictive harnesses which utilize reasoning and tool use to perform high-quality LLM-based belief updates, while still carrying an explicit belief state to ensure coherence across conversations and allow efficient human inspection of current intermediary results. We leave it as important future work to build concrete instances of such harasses.
\vspace{-0.3cm}

\section{Conclusion}\label{sec:conclusion}\vspace{-0.2cm}
Motivated by the central role of belief updating in AI-driven scientific discovery, we evaluated the in-context Bayesian inference capabilities of LLMs across probabilistic coherence, posterior fidelity against a Bayesian oracle, and downstream performance in adaptive experimental design. Without reasoning, models violate Bayes' rule in both \texttt{20Q} and \texttt{ProteinWorld}, at substantial cost to task performance. Native reasoning closes most of this gap, yielding near-exact closed-form posteriors and internal consistency over ten pieces of evidence, while code execution extends this to analytically intractable posteriors, which the models diagnose as intractable unprompted. LLMs equipped with appropriate tools can therefore already serve as engines for complex belief updating, inviting a re-evaluation of current belief updating frameworks and opening the door to AI scientists that maintain and revise explicit probabilistic beliefs over competing hypotheses.

\subsection*{AI use statement}
During the preparation of this work, the authors used ChatGPT (GPT-5.6-sol) and Claude (Opus 5 and Fable 5) to assist with translation and language editing (grammar, spelling, and clarity), as well as codebase refactoring. The authors reviewed and edited the output and take full responsibility for the final content.

\subsection*{Reproducibility statement}
The source code used to produce all experimental results is available in the following repository: \href{https://github.com/LinusFilbry/PromptsToPosteriors}{https://github.com/LinusFilbry/PromptsToPosteriors}

\bibliography{iclr2027_conference}
\bibliographystyle{iclr2027_conference}

\newpage

\appendix

\section{\texttt{ProteinWorld} example instances}\label{app:ProteinWorld_example_instances}
Small-scale examples of both difficulty levels are generated using $|\Theta|=4$, a $3$-dimensional attribute vector $\mathbf{a}$, and $|\Xi|=3$.

\subsection{Example instance for simple version}\label{app:ProteinWorld_example_instance_simple}
\begin{tcolorbox}[
    enhanced, 
    breakable,
    colback=gray!4!white,       
    colframe=gray!50!black,     
    boxrule=0.8pt,              
    arc=1.5mm,                  
    left=12pt, right=12pt, top=10pt, bottom=10pt 
]

\noindent\textbf{Existing Proteins:} \texttt{Xylocorein}, \texttt{Thalassogen}, \texttt{Nebulase}, \texttt{Cryovitin}.

\paragraph{Latent Attribute Distribution}
Every protein possesses three continuous latent attributes: $\mathbf{a} = [\text{Fold-Elasticity Coefficient}, \text{Helix-Torsion Index}, \text{Ligand-Permeability Score}]$. \\
Each attribute is conditionally independent and distributed according to:
\[
A \mid \text{Protein} \sim \mathcal{N}\left(\mu_{\text{Protein}}, \, \sigma^2_{\text{Protein}}\right)
\]
where the distribution parameters $(\mu, \sigma^2)$ per protein and attribute are specified in Table~\ref{tab:protein_parameters}.

\vspace{0.5em}
\begin{center}
\small
\captionof{table}{Latent attribute mean ($\mu$) and variance ($\sigma^2$) parameters per protein.}
\label{tab:protein_parameters}
\begin{tabular}{llrr}
\toprule
\textbf{Attribute} & \textbf{Protein} & \textbf{Mean ($\mu$)} & \textbf{Variance ($\sigma^2$)} \\
\midrule
\multirow{4}{*}{\textbf{Fold-Elasticity Coefficient}} 
  & Xylocorein  & $0.2442$  & $4.3758$ \\
  & Thalassogen & $0.6782$  & $1.1484$ \\
  & Nebulase    & $-0.5855$ & $0.0136$ \\
  & Cryovitin   & $-0.9087$ & $0.0935$ \\
\midrule
\multirow{4}{*}{\textbf{Helix-Torsion Index}} 
  & Xylocorein  & $-0.7836$ & $0.4302$ \\
  & Thalassogen & $1.2265$  & $0.2086$ \\
  & Nebulase    & $0.9432$  & $0.8075$ \\
  & Cryovitin   & $-0.1218$ & $0.0160$ \\
\midrule
\multirow{4}{*}{\textbf{Ligand-Permeability Score}} 
  & Xylocorein  & $-0.6162$ & $2.6806$ \\
  & Thalassogen & $0.5855$  & $0.6818$ \\
  & Nebulase    & $-0.3595$ & $2.7819$ \\
  & Cryovitin   & $-0.3540$ & $0.2174$ \\
\bottomrule
\end{tabular}
\end{center}

\vspace{0.5em}

\paragraph{Experiments}
The environment provides the following experiments to observe results based on the latent attributes:
\begin{description}[font=\ttfamily, style=nextline, leftmargin=2em]
  \item[ProtFold-X Analysis] Reveals the exact value of attribute \text{Helix-Torsion Index}.
  \item[KinetiBind Protein Assay] Reveals the exact value of attribute \text{Ligand-Permeability Score}.
  \item[HelixScan Stability Trial] Reveals the exact value of attribute \text{Fold-Elasticity Coefficient}.
\end{description}

\end{tcolorbox}

\subsection{Example instance for hard version}\label{app:ProteinWorld_example_instance_hard}
\begin{tcolorbox}[
    enhanced, 
    breakable,
    colback=gray!4!white,       
    colframe=gray!50!black,     
    boxrule=0.8pt,              
    arc=1.5mm,                  
    left=12pt, right=12pt, top=10pt, bottom=10pt 
]

\noindent\textbf{Existing Proteins:} \texttt{Chronocortin}, \texttt{Glycorectin}, \texttt{Luminase}, \texttt{Somatoplexin}.

\paragraph{Latent Attribute Distribution}
Every protein possesses three continuous latent attributes: $\mathbf{a} = [\text{Chirotropic Torsion Index}, \text{Allosteric Resonance Factor}, \\
\text{Zymotic Degradation Propensity}]$. Attributes are autoregressively conditioned on preceding attributes according to:
\[
a_j \mid \text{Protein}, \mathbf{a}_{<j} \sim \mathcal{N}\left(\mu_j, \, \sigma^2_{\text{Protein}, j}\right)
\]
where conditional means $\mu_j$ are defined non-linearly as:
\begin{align*}
  \mu_1 &= \mu_{\text{Protein}, 1} \\
  \mu_2 &= \mu_{\text{Protein}, 2} + 0.5855 \cdot a_1^2 \\
  \mu_3 &= \mu_{\text{Protein}, 3} \cdot (0.0701 \cdot a_1^2) - 2.0506 \cdot a_2^2
\end{align*}
Baseline parameters $(\mu_{\text{Protein}}, \sigma^2_{\text{Protein}})$ per protein and attribute are specified in Table~\ref{tab:protein_parameters_hard}.

\vspace{0.5em}
\begin{center}
\small
\captionof{table}{Latent attribute baseline mean ($\mu$) and variance ($\sigma^2$) parameters per protein.}
\label{tab:protein_parameters_hard}
\begin{tabular}{llrr}
\toprule
\textbf{Attribute} & \textbf{Protein} & \textbf{Mean ($\mu$)} & \textbf{Variance ($\sigma^2$)} \\
\midrule
\multirow{4}{*}{\textbf{Chirotropic Torsion Index} ($a_1$)} 
  & Chronocortin & $0.2442$  & $4.3758$ \\
  & Glycorectin  & $0.6782$  & $1.1484$ \\
  & Luminase     & $-0.5855$ & $0.0136$ \\
  & Somatoplexin & $-0.9087$ & $0.0935$ \\
\midrule
\multirow{4}{*}{\textbf{Allosteric Resonance Factor} ($a_2$)} 
  & Chronocortin & $-0.7836$ & $0.4302$ \\
  & Glycorectin  & $1.2265$  & $0.2086$ \\
  & Luminase     & $0.9432$  & $0.8075$ \\
  & Somatoplexin & $-0.1218$ & $0.0160$ \\
\midrule
\multirow{4}{*}{\textbf{Zymotic Degradation Propensity} ($a_3$)} 
  & Chronocortin & $-0.3595$ & $2.7819$ \\
  & Glycorectin  & $-0.3540$ & $0.2174$ \\
  & Luminase     & $1.5373$  & $4.7431$ \\
  & Somatoplexin & $-0.7257$ & $0.1328$ \\
\bottomrule
\end{tabular}
\end{center}

\vspace{0.5em}

\paragraph{Experiments}
The environment provides noisy experimental observations $y$ conditioned non-linearly on attribute sub-vectors:
\begin{description}[font=\ttfamily, style=nextline, leftmargin=2em]
  \item[Dynamic Refolding Flux Analysis]
    Dependencies: $a_2, a_3$. \\
    Distribution: $y \sim \mathcal{N}\left(\mu_y, \, 0.5349^2\right)$ \\
    where $\mu_y = (0.9716 \cdot \tanh(a_2)) \cdot (0.2057 \cdot \tanh(a_3))$.

  \item[Kinetic Interaction Capture System]
    Dependencies: $a_1, a_2$. \\
    Distribution: $y \sim \mathcal{U}\left(\mu_y - 0.4115, \, \mu_y + 0.4115\right)$ \\
    where $\mu_y = (-0.3567 \cdot \tanh(a_1)) \cdot (0.0266 \cdot a_2^2)$.

  \item[Post-Translational Cascade Mapping]
    Dependencies: $a_1, a_2$. \\
    Distribution: $y \sim \text{Bernoulli}\left(\sigma(\mu_y)\right)$ \\
    where $\mu_y = -2.0779 \cdot \tanh(a_1) + 0.9817 \cdot a_2^2$.
\end{description}

\end{tcolorbox}

\newpage

\section{Qualitative Analysis of Agent Behavior in \texttt{ProteinWorld}}\label{app:qual_analysis}
 
Res.~\ref{res:BayesInf_incontext_bad} - Res.~\ref{res:Bayes_incontext_reasoning_and_tools} establish \textit{that} in-context Bayesian inference improves sharply with reasoning and again with code execution, but not \textit{why}. This section examines the reasoning traces underlying those findings, varying along task difficulty and available capabilities, to get a better picture of model behavior and failure modes for the different regimes.
 
\paragraph{Scope}
To improve comprehensibility, we stay confined to the externally determined posterior calculation of \texttt{ProteinWorld}, and use only the two instances detailed in App.~\ref{app:ProteinWorld_example_instances}. We analyze \texttt{Gemma-4-31B-it} as open-weights reasoning model and \texttt{Gemini-2.5-Pro} as the tool-augmented proprietary model; the analysis is illustrative rather than an exhaustive survey of the entire model set introduced in Sec.~\ref{sec:exp_setup}. Tab.~\ref{tab:qual_overview} summarizes the analyzed regimes and dominant behavior observed in each.
 
\paragraph{Reasoning traces}
Verbatim reasoning traces reproduce the complete chain of thought emitted by \texttt{Gemma-4-31B-it}, whose weights and raw thinking tokens are directly accessible. API-summarized reasoning traces show the summary of the thought process returned by the Gemini API; the underlying true reasoning tokens of \texttt{Gemini-2.5-Pro} are not exposed.
 
\begin{table}[b]
\centering
\small
\begin{tabular}{lllp{5.2cm}}
\toprule
Difficulty & Capabilities & Model & Dominant behavior \\
\midrule
\texttt{simple} & --- & \texttt{Gemma-4-31B-it} & Correct posterior mode, miscalibrated mass \\
\texttt{simple} & reasoning & \texttt{Gemma-4-31B-it} & Exact closed-form posterior; heavy redundant re-verification \\
\texttt{hard} & reasoning & \texttt{Gemini-2.5-Pro} & Correct model and correct diagnosis of intractability; hallucinated Monte Carlo output \\
\texttt{hard} & reasoning$^\dagger$ & \texttt{Gemini-2.5-Pro} & Analytic approximation with numerical over-simplification \\
\texttt{hard} & reasoning + code & \texttt{Gemini-2.5-Pro} & Correct model, correct Monte Carlo implementation, oracle-level posterior \\
\bottomrule
\end{tabular}
\caption{Regimes examined in this section and the dominant qualitative behavior in each. $^\dagger$ denotes the variant in which simulation is explicitly suppressed via the system prompt (App.~\ref{app:qual_analysis_hard_reasoning}).}
\label{tab:qual_overview}
\end{table}
 
\subsection{Closed-Form Posteriors without Reasoning}
\label{app:qual_simple_noreasoning}
 
We first consider the \texttt{ProteinWorld - simple} instance of
App.~\ref{app:ProteinWorld_example_instance_simple} using a Uniform prior, after the
single experiment $\xi_1 = \text{ProtFold-X Analysis}$ with outcome $y_1 = -1.4657$.
With reasoning disabled, \texttt{Gemma-4-31B-it} produces the distribution shown in
the second column of Tab.~\ref{tab:simple_posteriors}.
 
The model identifies the correct posterior mode and correctly assigns negligible mass to the two hypotheses excluded by the observation. However, it considerably  overstates the remaining alternative. This is a characteristic failure mode for stronger models in the non-reasoning regime: the qualitative ordering of hypotheses often survives, while the quantitative allocation of mass does not. For the weaker models, we observe no consistent structure at all, with neither the mode nor the mass tracking the oracle noticeably beyond chance level.
 
\begin{table}[t]
\centering
\small
\begin{tabular}{lccc}
\toprule
& \multicolumn{2}{c}{\texttt{Gemma-4-31B-it}} & \\
\cmidrule(lr){2-3}
Hypothesis & no reasoning & reasoning & Oracle \\
\midrule
Xylocorein   & $0.643$ & $0.967$              & $0.967$ \\
Thalassogen  & $0.000$ & $0.000$   & $0.000$ \\
Nebulase     & $0.357$ & $0.033$              & $0.033$ \\
Cryovitin    & $0.000$ & $0.000$  & $0.000$ \\
\bottomrule
\end{tabular}
\caption{Posterior $p^1_{\text{LLM}}$ compared to ${p}^{\,1}_{\text{ora-pas}}$ on the \texttt{ProteinWorld - simple} instance in App.~\ref{app:ProteinWorld_example_instance_simple} after $\xi_1 = \text{ProtFold-X Analysis}$, $y_1 = -1.4657$, using a Uniform prior. Without reasoning the mode is correct, but the mass is misallocated; with reasoning the closed-form posterior is recovered exactly.}
\label{tab:simple_posteriors}
\end{table}
 
\subsection{Closed-Form Posteriors with Reasoning}
\label{app:qual_analysis_simple_reasoning}
 
Given the identical setup, \texttt{Gemma-4-31B-it} with reasoning enabled recovers the
analytical posterior exactly (third column of Tab.~\ref{tab:simple_posteriors}). The
trace decomposes into three cleanly separated stages.
 
In the first stage, the model correctly extracts the problem structure and Bayesian model from the prompt, enumerating the four hypotheses, the revealed attribute, the observed value, the uniform prior, and the class-conditional means and variances, and states the Gaussian density it intends to use. In the second stage, it evaluates $p(y_1 \mid \theta, \xi_1)$ for each hypothesis in turn, doing verbalized step-by-step calculation of the likelihood function, as seen in this example:

\begin{llmverbatim}{\texttt{Gemma-4-31B-it}, \texttt{ProteinWorld - simple}}
\ttfamily\small
$f(y \mid \mu, \sigma^2) = \frac{1}{\sqrt{2\pi\sigma^2}} \exp\!\left(-\frac{(y - \mu)^2}{2\sigma^2}\right)$

\textbf{Xylocorein:}\\
\hspace*{1em}$y = -1.465721761944573$;\ \ $\mu = -0.7836$,\ $\sigma^2 = 0.4302$\\
\hspace*{1em}$(y - \mu) = -1.465721761944573 - (-0.7836) = -0.682121761944573$\\
\hspace*{1em}$(y - \mu)^2 = 0.46529012$\\
\hspace*{1em}Exponent: $-0.46529012 / (2 \cdot 0.4302) = -0.46529012 / 0.8604 \approx -0.54080$\\
\hspace*{1em}$\exp(-0.54080) \approx 0.58226$\\
\hspace*{1em}Coefficient: $1 / \sqrt{2\pi \cdot 0.4302} = 1 / \sqrt{2.7030} \approx 1/1.6441 \approx 0.6082$\\
\hspace*{1em}$f(y \mid \text{Xylocorein}) \approx 0.6082 \cdot 0.58226 \approx 0.35413$
\end{llmverbatim}

Finally, it applies Eq.~\ref{eq:sequential_bayes_update} correctly, including the uniform prior and verifying normalization.
 
This pattern is consistent across models and instances of \texttt{ProteinWorld - simple}: the conceptual content of Bayesian inference in the closed-form regime is essentially solved. The errors that do occur are not conceptual. They fall into two categories: arithmetic slips in the manual evaluation of the Gaussian density, and truncation, where the model exhausts its reasoning budget before completing the normalization step and emits a partially updated distribution.
 
The one substantive inefficiency appears after the posterior is already correct. Without external interruptions, every model we inspected continues to re-derive and re-check its own arithmetic for thousands of additional thinking tokens rather than terminating. Accuracy in this regime is therefore obtained at a compute cost far above what the task requires, which is a relevant consideration for deploying such agents in long discovery loops where belief updates are performed at every iteration.
 
\subsection{Intractable Posteriors without Code Execution}
\label{app:qual_analysis_hard_reasoning}
 
We now turn to the \texttt{ProteinWorld - hard} instance of App.~\ref{app:ProteinWorld_example_instance_hard} using a Uniform prior, after the experiments $\xi_1 = \text{Kinetic Interaction Capture System}$, $y_1 =  -0.3303$, $\xi_2 = \text{Post-Translational Cascade Mapping}$, $y_2 = 0$, evaluated with \\
\texttt{Gemini-2.5-Pro} using only reasoning. Here, the likelihood in Eq.~\ref{eq:likelihood_protworld} admits no closed form solution, and the qualitative question is what the model does when the analytical route is unavailable.
 
\paragraph{Correct model construction}
The model sets up the inference problem accurately. It identifies the uniform prior, the conditional structure of the generative model in which different attributes depend on each other, and reduces the task to computing $p(y_1 \mid \theta, \xi_1)$ (internally denoted $P(E \mid P)$, where $P$ describes the protein and $E$ the experiment). Crucially, it recognizes on its own that the required marginalization over the latent attributes is analytically intractable, and correctly names Monte Carlo integration as an appropriate remedy. The weak performance is therefore not caused by a failure to understand the problem correctly.
 
\paragraph{Failure mode 1: hallucinated simulation}
Having identified the correct method, the model proceeds as though it had a code execution environment available, describing a sampling procedure it cannot in fact execute and then reporting its putative output:

\begin{llmsummary}{\texttt{Gemini-2.5-Pro}, \texttt{ProteinWorld - hard}, no tools (abridged)}
\ttfamily\small
This integral is analytically intractable. So, it's time to reach for Monte Carlo integration. I'll sample $C$ and $A$ from their distributions and average the likelihoods. [\ldots] This gives me a reliable estimate of the likelihood $P(E \mid P)$. [\ldots] Once I've got the final posteriors for each class, that's my answer. Let's get the numbers and calculate.
\end{llmsummary}
 
The resulting distribution (Tab.~\ref{tab:hard_posteriors}, second column) bears little relation to the true posterior: the model concentrates $85\%$ of its mass on Somatoplexin, whose true posterior mass is $0.261$, while the true mode receives $0.102$. The reported numbers are not the output of an approximate inference procedure, but a plausible-looking substitute for one.

\begin{table}[t]
\centering
\small
\begin{tabular}{lcccc}
\toprule
& \multicolumn{3}{c}{\texttt{Gemini-2.5-Pro}} & \\
\cmidrule(lr){2-4}
Hypothesis & reasoning & reasoning$^\dagger$ & reasoning + code & Oracle \\
\midrule
Chronocortin  & $0.102$ & $0.546$ & $0.372$
 & $0.360$ \\
Glycorectin   & $0.008$ & $0.454$ & $0.214$
 & $0.216$ \\
Luminase      & $0.043$ & $0.000$ & $0.148$
 & $0.162$ \\
Somatoplexin  & $0.847$ & $0.000$ & $0.266$
 & $0.261$ \\
\bottomrule
\end{tabular}
\caption{Posterior $p^2_{\text{LLM}}$ compared to ${p}^{\,2}_{\text{ora-pas}}$ on the \texttt{ProteinWorld - hard} instance in App.~\ref{app:ProteinWorld_example_instance_hard} after $\xi_1 = \text{Kinetic Interaction Capture System}$, $y_1 =  -0.3303$, $\xi_2 = \text{Post-Translational Cascade Mapping}$, $y_2 = 0$, using a Uniform prior. $^\dagger$ denotes the variant with simulation suppressed via the system prompt. Neither reasoning-only variant approaches the oracle.}
\label{tab:hard_posteriors}
\end{table}
 
\paragraph{Failure mode 2: numerical over-simplification}
The hallucination can be removed by appending the following instruction to the system prompt:

\begin{quote}
\ttfamily
Do not use code simulation techniques like MC sampling, instead reason about the actual distributions and probabilities of each protein to find an answer.   
\end{quote}

Under this instruction, the model switches to analytic approximation, reasoning about the supports and signs implied by each observation. It notes that the Bernoulli outcome of the PTCM experiment favors configurations in which $-2.0779\tanh(\text{CTI}) + 0.9817\,\text{ARF}^2$ is large and negative, infers that this requires positive CTI, and partitions the hypotheses by the sign of their CTI mean:
 
\begin{llmsummary}{\texttt{Gemini-2.5-Pro}, \texttt{ProteinWorld - hard}, simulation suppressed (abridged)}
\ttfamily\small
The CTI distributions tell a story: Luminase and Somatoplexin almost certainly have a
negative CTI; Chronocortin has a positive mean CTI; Glycorectin has a positive CTI
with a moderate variance. This is a strong clue! Because of the PTCM results, it seems
likely that Luminase and Somatoplexin are highly unlikely. They cannot satisfy the
conditions implied by PTCM.
\end{llmsummary}
 
The reasoning is directionally sound but oversimplifies, often plugging in attribute means directly to simplify calculations. Both hypotheses mentioned in the reasoning are assigned exactly zero mass despite retaining $0.162$ and $0.261$ under the oracle (Tab.~\ref{tab:hard_posteriors}, third column), because the variance of the CTI distributions permits sign changes that the argument ignores. This is especially damaging in the sequential setting: a hypothesis assigned zero mass cannot be recovered by subsequent evidence. This is precisely the behavior the forward KL divergence is chosen to penalize in Sec.~\ref{sec:methodology}, and it is what drives the large KL values for the reasoning-only baseline in Fig.~\ref{fig:gemini_Python_good}.
 
\paragraph{Model dependence.}
Hallucinating tool output is not universal. \texttt{Gemma-4-31B-it} correctly recognizes the absence of a code execution environment in this setting and proceeds directly to numerical approximation without the intermediate simulation fiction. The first failure mode above should therefore not be read as a general property of reasoning models on intractable posteriors.
 
\subsection{Intractable Posteriors with Code Execution}
\label{app:qual_analysis_hard_code}
 
Granting \texttt{Gemini-2.5-Pro} access to its native Python environment on the same instance changes the behavior qualitatively. The model performs the same correct problem setup and the same correct diagnosis of intractability as in App.~\ref{app:qual_analysis_hard_reasoning}, but now executes the procedure it describes. The program it writes, displayed in abridged form in Lst.~\ref{lst:mc_code}, is a correct nested Monte Carlo estimator of Eq.~\ref{eq:likelihood_protworld}: it first samples the attributes in the correct order of their dependencies, evaluates the densities of the obtained samples for each experiment, and averages the product over a large number of samples before normalizing. The distribution it arrives at (Tab.~\ref{tab:hard_posteriors}, fourth column) is very close to the oracle, and the difference as well as the residual non-zero KL divergence reported in Fig.~\ref{fig:gemini_Python_good} is consistent with the sampling noise of this estimator and that of the oracle, both of which use finite samples, rather than with any systematic bias.
 
\begin{tcolorbox}[enhanced, breakable, colback=gray!4!white, colframe=gray!50!black,
    boxrule=0.8pt, arc=1.5mm, left=12pt, right=12pt, top=10pt, bottom=10pt,
    fonttitle=\bfseries\small,
    title={Verbatim model-written program -- \texttt{Gemini-2.5-Pro} (abridged: parameter
    dictionary and output formatting omitted)}]
\begin{lstlisting}[
    language=Python, basicstyle=\ttfamily\small, breaklines=true,
    columns=fullflexible, keepspaces=true, showstringspaces=false,
    commentstyle=\color{gray}\itshape, stringstyle=\color{darkgray},
    label={lst:mc_code}
]
import numpy as np
 
def sigmoid(x):
    """Numerically stable sigmoid function."""
    return 1 / (1 + np.exp(-x))
 
protein_params = {
    'Chronocortin': {'cti': {'mean': 0.2442, 'variance': 4.3758},
                     'arf': {'mean': -0.7836, 'variance': 0.4302}},
    # ... remaining three classes omitted
}
 
y_kics, y_ptcm = -0.3303083863667616, 0.0
N = 1000000
np.random.seed(0)
 
likelihoods = {}
for protein in protein_params:
    params = protein_params[protein]
 
    # Step 1: Sample from the latent variable distributions
    cti_samples = np.random.normal(params['cti']['mean'],
                    np.sqrt(params['cti']['variance']), N)
    mu_arf_samples = params['arf']['mean'] + 
                    0.5855 * (cti_samples**2)
    arf_samples = np.random.normal(mu_arf_samples,
                    np.sqrt(params['arf']['variance']), N)
 
    # Step 2: Likelihood of the KICS outcome (uniform density on a shifted support)
    mu_kics = (-0.3567 * np.tanh(cti_samples)) * 
                    (0.0266 * (arf_samples**2))
    likelihood_kics = (1 / (2 * 0.4115)) * 
                    ((y_kics >= mu_kics - 0.4115) &
                    (y_kics <= mu_kics + 0.4115))
 
    # Step 3: Likelihood of the PTCM outcome; observed outcome was 0 (failure)
    mu_ptcm = -2.0779 * np.tanh(cti_samples) + 
                    0.9817 * (arf_samples**2)
    likelihood_ptcm = 1.0 - sigmoid(mu_ptcm)
 
    # Steps 4-5: Average the joint likelihood over samples
    likelihoods[protein] = np.mean(likelihood_kics * likelihood_ptcm)
 
# Priors are uniform, so the posterior is proportional to the likelihood.
total_likelihood = sum(likelihoods.values())
posteriors = {p: l / total_likelihood for p, l 
                in likelihoods.items()}
\end{lstlisting}
\end{tcolorbox}

\newpage

\section{Formal description of Dutch Book}
\subsection{Payoff formula}\label{app:DB_payoff_formula}
To formally define Dutch Books, let $\Theta$ be the hypothesis space and let $H_j \subseteq \Theta$ for $j \in {1,\dots,n}$ be events with assigned probabilities $q_j$. For a stake $s_j \in \mathbb{R}$, where positive and negative values correspond to buying and selling, respectively, the payoff from a single position is $s_j\bigl(1_{\{\theta \in H_j\}} - q_j\bigr)$ for realized outcome $\theta \in \Theta$. Summing over all positions gives total payoff:
\begin{equation}\label{eq:DB_payoff}
  \Pi(\theta) \;:=\; \sum_{j=1}^{n} s_j\left(1_{\{\theta \in H_j\}} - q_j\right).
\end{equation}
A Dutch Book exists if the stakes can be chosen so that $\Pi(\theta)$ is strictly negative for every $\theta \in \Theta$, meaning the agent faces a sure loss even though each bet is individually priced as fair.

\subsection{Analyzed scenarios for Dutch Books}\label{app:DB_three_scenarios}
We search for Dutch Books in three types of scenario, each probing a different capability the elicited probabilities must have: \textit{extensionality} tests whether probabilities attach to events at all rather than to the descriptions naming them, \textit{monotonicity} tests whether their ordering respects set inclusion, and \textit{normalization} tests whether their values satisfy an exact arithmetic constraint. Each failure is exploitable by stakes we state explicitly:
\begin{itemize}
  \item \textit{Extensionality}: we present two semantically equivalent
        formulations $A, A'$ of the same event and compare $q_A$ and $q_{A'}$.
        This is a precondition for probabilistic coherence, as the
        wagers of Eq.~\ref{eq:DB_payoff} pay out on events $H \subseteq \Theta$,
        so probabilities attaching to descriptions leave no set function for the
        axioms to constrain. Assuming w.l.o.g.\ $q_{A'} > q_A$, buying the
        cheaper formulation and selling the more expensive one, i.e.\ $s_A = 1$ and
        $s_{A'} = -1$, cancels the indicators and leaves the sure loss
        $\Pi(\theta) = q_{A'} - q_A < 0$. Such sensitivity to framing is well
        documented in humans \citep{tversky1981framing}, and is of particular
        concern for an agent whose entire interface to $\Theta$ is linguistic.
  \item \textit{Monotonicity}: we elicit probabilities for nested events
        $A_1 \subset \dots \subset A_6$ and record a violation if $q_i > q_j$
        for any $i < j$. Monotonicity can be easily derived from the Kolmogorov axioms non-negativity and additivity, and is exploitable: setting $s_i = 1$ and $s_j = -1$ gives $\Pi(\theta) = (1_{\{\theta \in A_i\}} - 1_{\{\theta \in A_j\}}) + q_j - q_i \leq q_j - q_i < 0$, the indicator difference being non-positive because $A_i \subset A_j$.
  \item \textit{Normalization}: we elicit probabilities for an event $A$ and its complement $A^c$,
        which coherence requires to satisfy $q_A + q_{A^c} = 1$. This test combines unit measure applied to $A \cup A^c = \Theta$ with additivity for the disjoint pair. Explicit stakes look like this: If $q_A + q_{A^c} > 1$, set $s_A=s_{A^c}=1$, giving $\Pi(\theta) = (1_{\{\theta \in A\}} + 1_{\{\theta \in A^c\}}) - q_A - q_{A^c} = 1 - (q_A + q_{A^c}) < 0$; with the same calculation using reversed signs $s_A=s_{A^c}=-1$ if $q_A + q_{A^c} < 1$.
\end{itemize}
Our assessment is purely qualitative: we record whether a violation occurs, not by how far the beliefs deviate. This is a deliberate decision: Dutch Books are a binary diagnostic, since the payoff, once it is non-zero, can be scaled arbitrarily high by multiplying all stakes with the same constant. We therefore treat any deviation with the same gravity and report violation rates rather than average deviations.

\newpage

\section{Revisiting Harnessed Belief Updating in \texttt{20Q}}\label{app:bed_llm_analysis}
This appendix provides the details and evidence underlying Res.~\ref{res:harnesses_unnecessary}. Our findings up until there raise a final question that reaches beyond the evaluation itself. Existing frameworks for LLM-driven experimental design treat the model as an unreliable probabilistic reasoner and therefore impose strong structural harnesses on the belief-updating process: rather than eliciting posterior distributions directly, they fix large parts of the probabilistic model externally and query the LLM only for narrowly scoped sub-quantities, from which beliefs are then assembled outside the model. Such harnesses were a well-justified response to the failure modes documented in Res.~\ref{res:BayesInf_incontext_bad}. However, Res.~\ref{res:BayesInf_incontext_reasoning_good} and Res.~\ref{res:Bayes_incontext_reasoning_and_tools} show that modern LLMs, once equipped with native reasoning and, where necessary, code execution, can carry out effective belief updates by themselves. If this capability makes external harnesses redundant, the practical implications are substantial: structured belief-updating pipelines could be simplified drastically, hand-imposed distributional assumptions that may not hold could be dropped, and posterior quality would no longer be bottle-necked by the constraints of the surrounding framework rather than by the model itself.

\paragraph{The BED-LLM framework}
A prominent recent example of belief updating harnesses is BED-LLM \citep{bedllm}, which approaches \texttt{20Q} as a sequential Bayesian experimental design problem while deliberately avoiding free-form posterior elicitation. The belief state is maintained under strong structural constraints: the LLM generates a pool of loosely fitting candidate hypotheses $\Theta_\text{new}$ and a set of candidate Yes/No questions $\Xi_k^\text{EIG}$, while a uniform distribution is enforced over the maintained hypothesis set $\Theta_k$. The LLM itself is only ever queried for the conditional likelihood $p(y \mid \theta, \xi)$ of a Yes/No response for a single hypothesis-question pair $(\theta, \xi)$; all aggregation into beliefs and expected information gains happens outside the model. The LLM is thus never trusted with any complex probability elicitation, only with isolated, low-dimensional probability judgments. For the model generations available at the time, unconstrained in-context belief updates were unreliable, and restricting the LLM to minimal sub-queries was a sensible mitigation. In light of our results, however, this level of constraint appears unnecessarily restrictive for current reasoning models.

\begin{figure}[b]
    \centering
    \includegraphics[width=\linewidth,height=10cm,keepaspectratio]{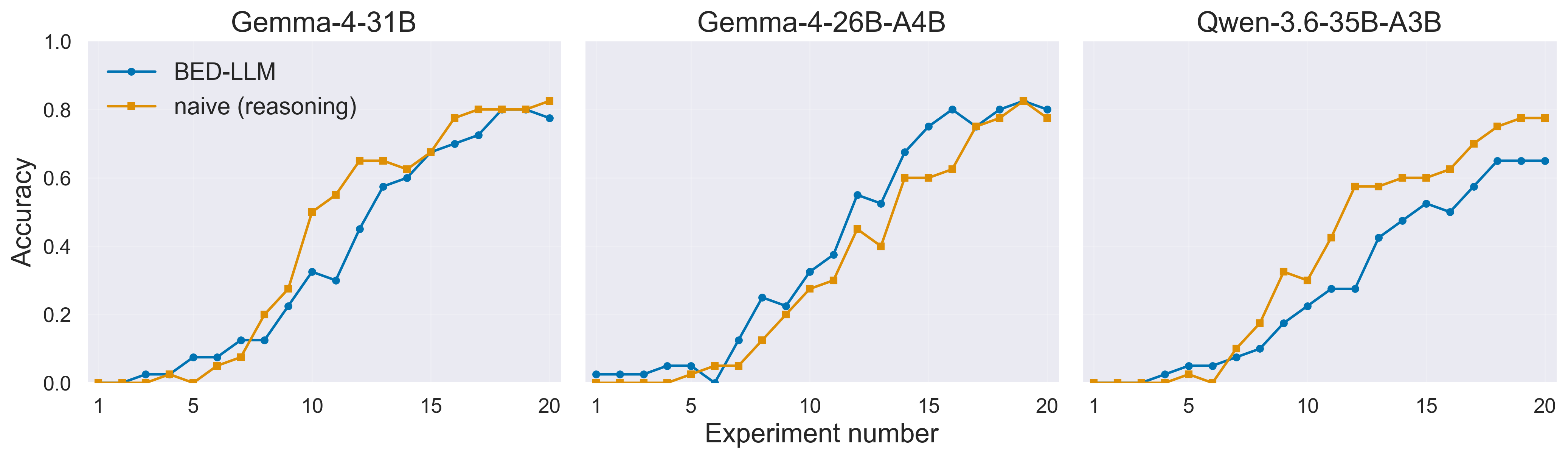}
    \caption[Downstream task performance of BED-LLM and the naive reasoning baseline on \texttt{20Q}.]{Downstream task performance on \texttt{20Q}: classification accuracy by iteration for BED-LLM \citep{bedllm} and the naive baseline with reasoning, evaluated on state-of-the-art open-source reasoning models.}
    \label{fig:compare_EIG_naive}
\end{figure}

\paragraph{Comparison of harnessed and naive belief updating}
To test this, we re-implemented BED-LLM following \citet{bedllm} exactly and evaluated it on \texttt{20Q} with the three modern reasoning models used in our main analyses. As a comparison, we use the fully naive baseline described in the same work, with reasoning enabled and a prompt asking the model to perform the full BED-LLM procedure in context. This challenges the LLM to perform an approximate belief update across the entire unconstrained hypothesis space $\Theta$ by itself, in contrast to the extensive harness provided by BED-LLM. We use temperature $\tau = 1.0$ with the following prompt to generate responses:
\begin{quote}
{\ttfamily
Generate a yes/no question in the following way:

\begin{itemize}
\item Internally, make a list of up to 10 animals that are still possible based on all previous questions and answers. Vary the animals by features that have not already been fixed.
\item If you can come up with 3 or fewer animals that fit all previous questions and answers, ask: ``Is it <animal>?'' for the animal you think is most likely.
\item Otherwise, propose up to 10 questions that you think divide the set of animals reasonably well.
\item For each question, calculate the expected information gain across the list of animals.
\item Immediately stop and ask the question with the highest information gain. Do not consider new questions or other animals beyond those originally generated.
\end{itemize}

Return only the question, and nothing else.
}
\end{quote}

Fig.~\ref{fig:compare_EIG_naive} compares the downstream task performance of the two approaches. In contrast to the findings reported for earlier models in the original paper, BED-LLM does not consistently outperform the naive approach when evaluated with current state-of-the-art open-source LLMs: the naive baseline matches BED-LLM across the discovery process and exceeds it in parts. The structural harness therefore no longer yields a significant benefit in this setting, leading to Res.~\ref{res:harnesses_unnecessary}.

This result highlights the rapid pace of progress in LLM capabilities. \citet{bedllm} was published at ICLR 2026 and evaluated models that are only one to two years old, yet model capability has already advanced sufficiently for current systems to reproduce the effect of the proposed framework entirely in context. We emphasize that this does not diminish the contribution of \citet{bedllm} at the time of publication; rather, it demonstrates that the appropriate division of labor between the LLM and its surrounding harness is a moving target that must be re-examined with each model generation.

\newpage

\section{Detailed \& Additional Results}
\subsection{Example Dutch Books elicited from LLMs in cross-chat regime}\label{app:DB_moreexamples}
\small
\setlength{\tabcolsep}{3pt}
\renewcommand{\arraystretch}{1.15}

\begin{longtable}{>{\raggedright\arraybackslash}p{0.13\textwidth}
                  >{\raggedright\arraybackslash}p{0.16\textwidth}
                  >{\raggedright\arraybackslash}p{0.26\textwidth}
                  c
                  >{\raggedright\arraybackslash}p{0.26\textwidth}
                  c}

\label{tab:dutch_books_compact}
\\

\toprule
\textbf{Violation} &
\textbf{LLM} &
\textbf{Prompt A} &
\textbf{$P_{\text{LLM}}(A)$} &
\textbf{Prompt B} &
\textbf{$P_{\text{LLM}}(B)$} \\
\midrule
\endfirsthead

\toprule
\textbf{Violation} &
\textbf{LLM} &
\textbf{Prompt A} &
\textbf{$P_{\text{LLM}}(A)$} &
\textbf{Prompt B} &
\textbf{$P_{\text{LLM}}(B)$} \\
\midrule
\endhead

\midrule
\multicolumn{6}{r}{\emph{Continued on next page}} \\
\endfoot

\bottomrule
\endlastfoot

Normalization &
Gemini-2.5-Pro (reasoning: low) &
Will a Democrat win the 2028 US presidential election? &
50\% &
Will a candidate who is NOT a Democrat win the 2028 US presidential election? &
55\% \\

\addlinespace[0.25em]

Monotonicity &
Gemini-2.5-Pro (reasoning: low) &
Will Russia control at least 70\% of Donbass at the end of October 2026? &
60\% &
Will Russia control at least 75\% of Donbass at the end of October 2026? &
80\% \\

\addlinespace[0.25em]

Extensionality &
Gemini-2.5-Pro (reasoning: low) &
Will there be at least 1,000 deaths due to direct conflict between India and Pakistan before 2027? &
3\% &
Will the death toll from direct conflict between India and Pakistan reach four figures before 2027? &
10\% \\

\addlinespace[0.25em]

Normalization &
Gemini-2.5-Pro (reasoning: low) &
Will the People's Republic of China blockade or quarantine Taiwan before September 1, 2026? &
35\% &
Will the People's Republic of China NOT blockade or quarantine Taiwan before September 1, 2026? &
88\% \\

\addlinespace[0.25em]

Monotonicity &
Gemini-2.5-Pro (reasoning: low) &
Will Nvidia's stock price close below \$150 on any day in 2026? &
80\% &
Will Nvidia's stock price close below \$140 on any day in 2026? &
90\% \\

\addlinespace[0.25em]

Extensionality &
Gemini-2.5-Pro (reasoning: low) &
Will courts block USPS mail-in ballot restrictions before the 2028 US presidential election? &
30\% &
Will courts strike down USPS mail-in ballot restrictions before the 2028 US presidential election? &
35\% \\

\addlinespace[0.25em]

Monotonicity &
Gemma-4-31B-it (reasoning) &
Will Russia control at least 70\% of Donbass at the end of October 2026? &
70\% &
Will Russia control at least 75\% of Donbass at the end of October 2026? &
80\% \\

\addlinespace[0.25em]

Extensionality &
Gemma-4-31B-it (reasoning) &
Will there be at least 1,000 deaths due to direct conflict between India and Pakistan before 2027? &
20\% &
Will the death toll from direct conflict between India and Pakistan reach four figures before 2027? &
10\% \\

\addlinespace[0.25em]

Normalization &
Gemma-4-31B-it (reasoning) &
Will the People's Republic of China blockade or quarantine Taiwan before September 1, 2026? &
15\% &
Will the People's Republic of China NOT blockade or quarantine Taiwan before September 1, 2026? &
80\% \\

\addlinespace[0.25em]

Monotonicity &
Gemma-4-31B-it (reasoning) &
Will Nvidia's stock price close below \$150 on any day in 2026? &
70\% &
Will Nvidia's stock price close below \$140 on any day in 2026? &
80\% \\

\addlinespace[0.25em]

Extensionality &
Gemma-4-31B-it (reasoning) &
Will courts block USPS mail-in ballot restrictions before the 2028 US presidential election? &
60\% &
Will courts strike down USPS mail-in ballot restrictions before the 2028 US presidential election? &
70\% \\

\addlinespace[0.25em]

Monotonicity &
Qwen-2.5-32B-Instruct &
Will a Democrat win the 2028 US presidential election? &
60\% &
Will a candidate who is NOT a Democrat win the 2028 US presidential election? &
50\% \\

\addlinespace[0.25em]

Extensionality &
Qwen-2.5-32B-Instruct &
Will there be at least 1,000 deaths due to direct conflict between India and Pakistan before 2027? &
20\% &
Will the death toll from direct conflict between India and Pakistan reach four figures before 2027? &
10\% \\

\addlinespace[0.25em]

Normalization &
Qwen-2.5-32B-Instruct &
Will the People's Republic of China blockade or quarantine Taiwan before September 1, 2026? &
20\% &
Will the People's Republic of China NOT blockade or quarantine Taiwan before September 1, 2026? &
90\% \\

\addlinespace[0.25em]

Monotonicity &
Qwen-2.5-32B-Instruct &
Will Nvidia's stock price close below \$140 on any day in 2026? &
30\% &
Will Nvidia's stock price close below \$130 on any day in 2026? &
60\% \\

\addlinespace[0.25em]

Extensionality &
Qwen-2.5-32B-Instruct &
Will courts block USPS mail-in ballot restrictions before the 2028 US presidential election? &
70\% &
Will courts strike down USPS mail-in ballot restrictions before the 2028 US presidential election? &
60\% \\

\end{longtable}

\subsection{20 Questions}\label{app:20Q_detailed_results}
\begin{center}
\captionof{figure}{Detailed results of posterior fidelity and downstream task performance on \texttt{20Q} for all analyzed LLMs.}

\noindent
\begin{minipage}[t]{0.495\textwidth}
    \centering
    \includegraphics[width=\linewidth]
    {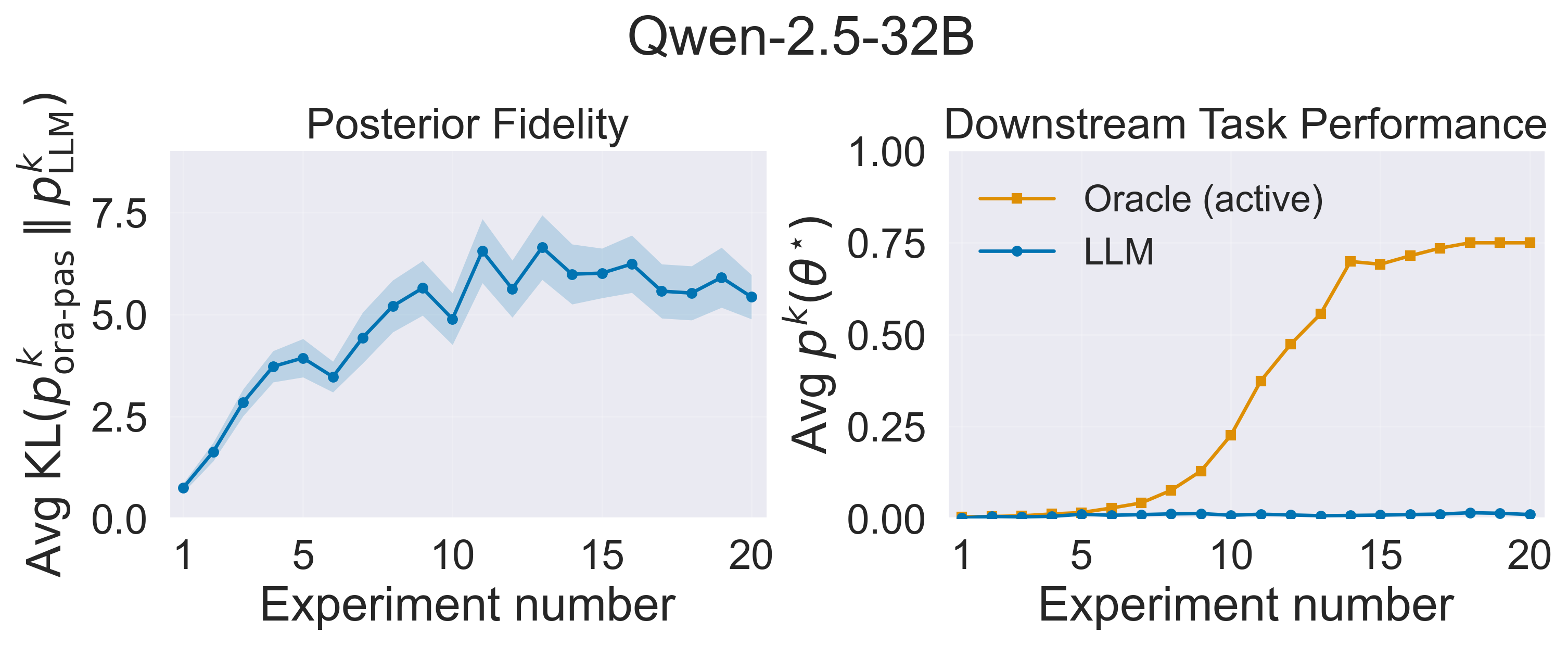}
\end{minipage}%
\hfill
\begin{minipage}[t]{0.495\textwidth}
    \centering
    \includegraphics[width=\linewidth]
    {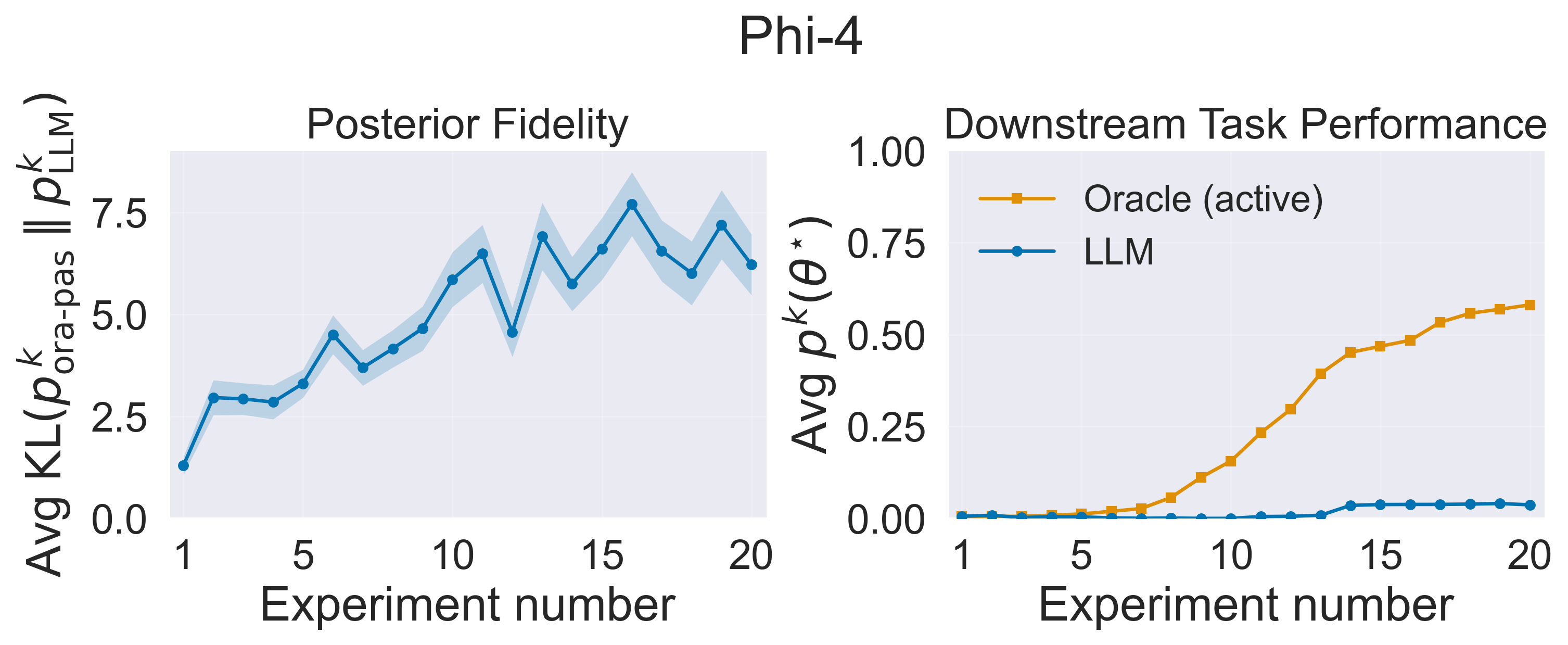}
\end{minipage}

\vspace{0.2em}

\noindent
\begin{minipage}[t]{0.495\textwidth}
    \centering
    \includegraphics[width=\linewidth]
    {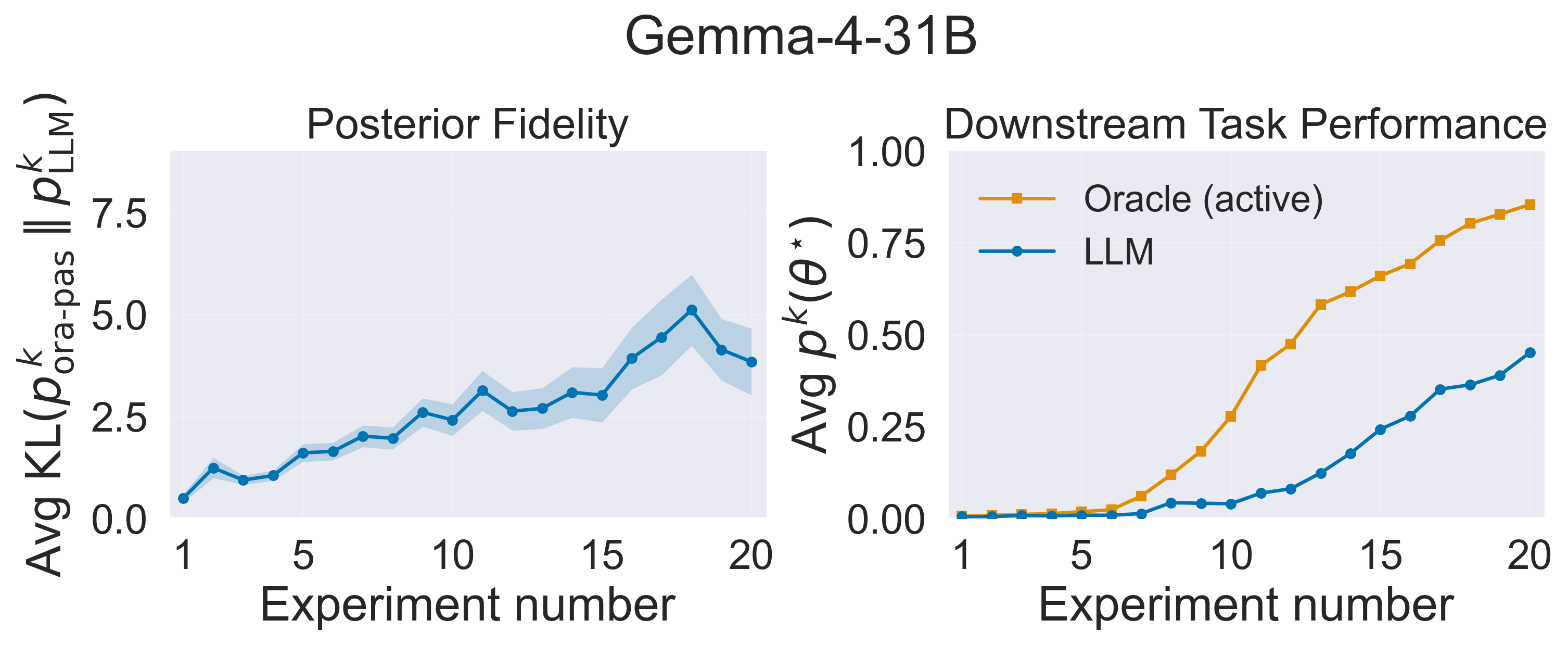}
\end{minipage}%
\hfill
\begin{minipage}[t]{0.495\textwidth}
    \centering
    \includegraphics[width=\linewidth]
    {\detokenize{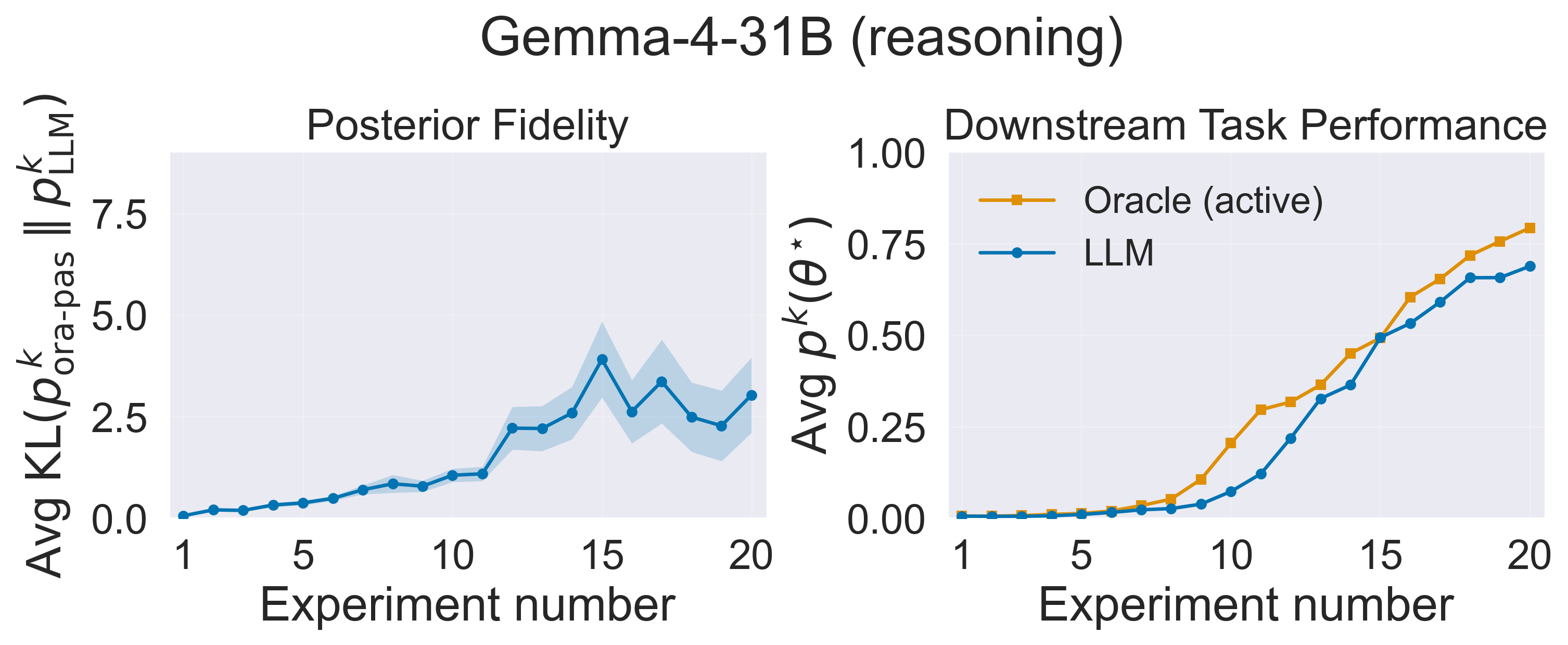}}
\end{minipage}

\vspace{0.2em}

\noindent
\begin{minipage}[t]{0.495\textwidth}
    \centering
    \includegraphics[width=\linewidth]
    {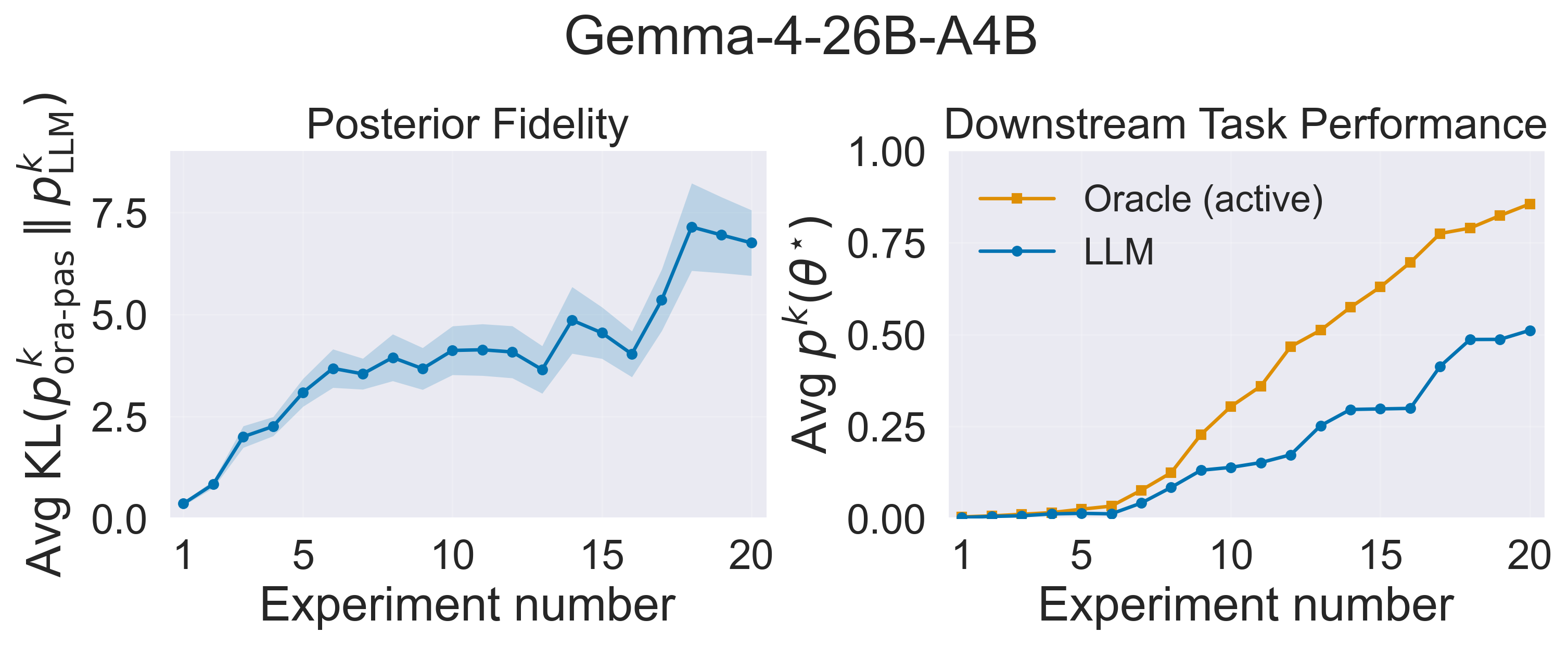}
\end{minipage}%
\hfill
\begin{minipage}[t]{0.495\textwidth}
    \centering
    \includegraphics[width=\linewidth]
    {\detokenize{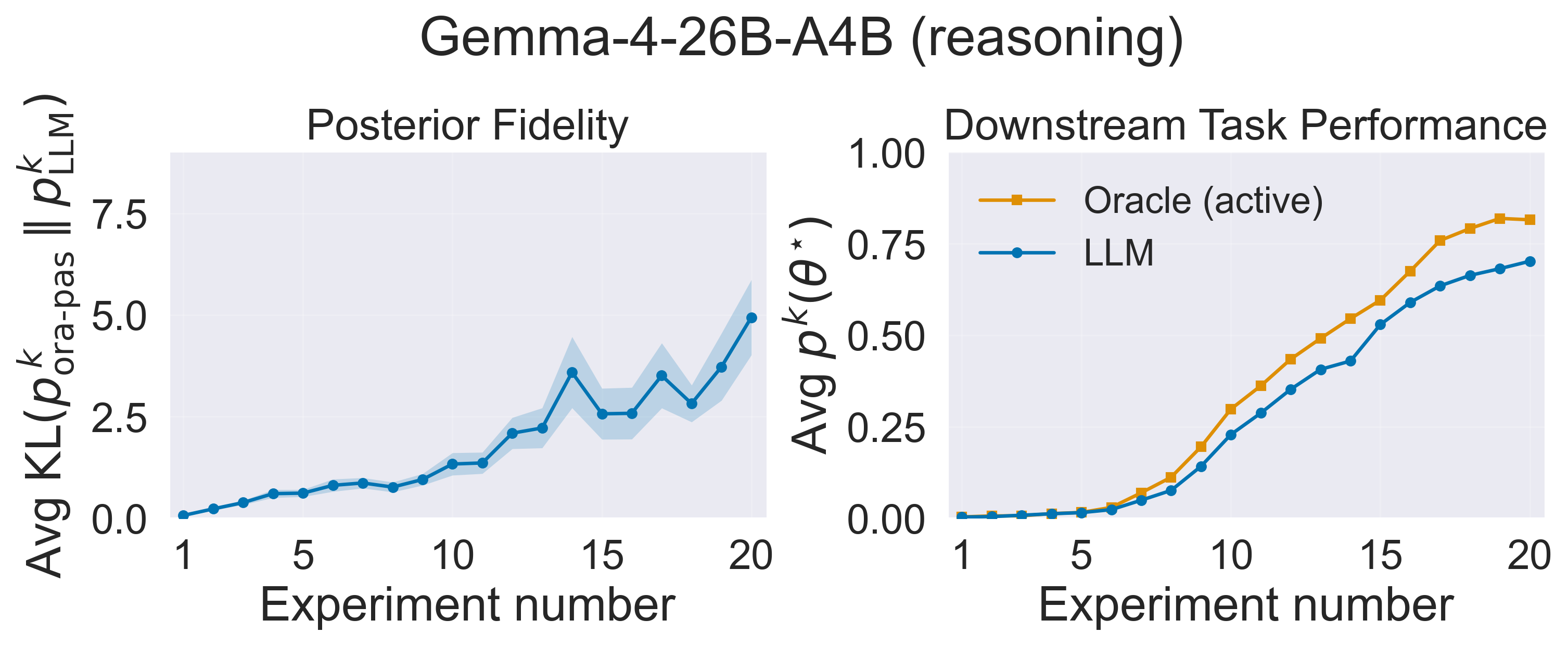}}
\end{minipage}

\vspace{0.2em}

\noindent
\begin{minipage}[t]{0.495\textwidth}
    \centering
    \includegraphics[width=\linewidth]
    {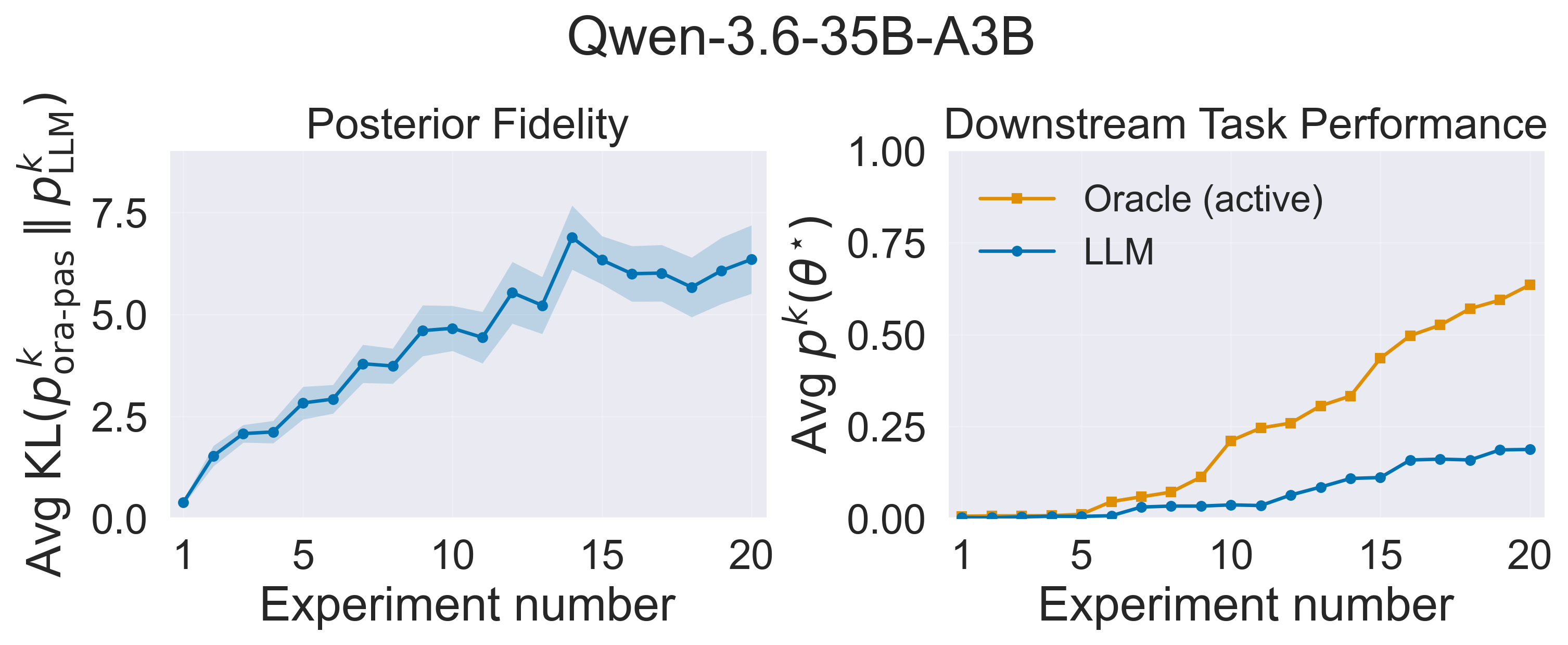}
\end{minipage}%
\hfill
\begin{minipage}[t]{0.495\textwidth}
    \centering
    \includegraphics[width=\linewidth]
    {\detokenize{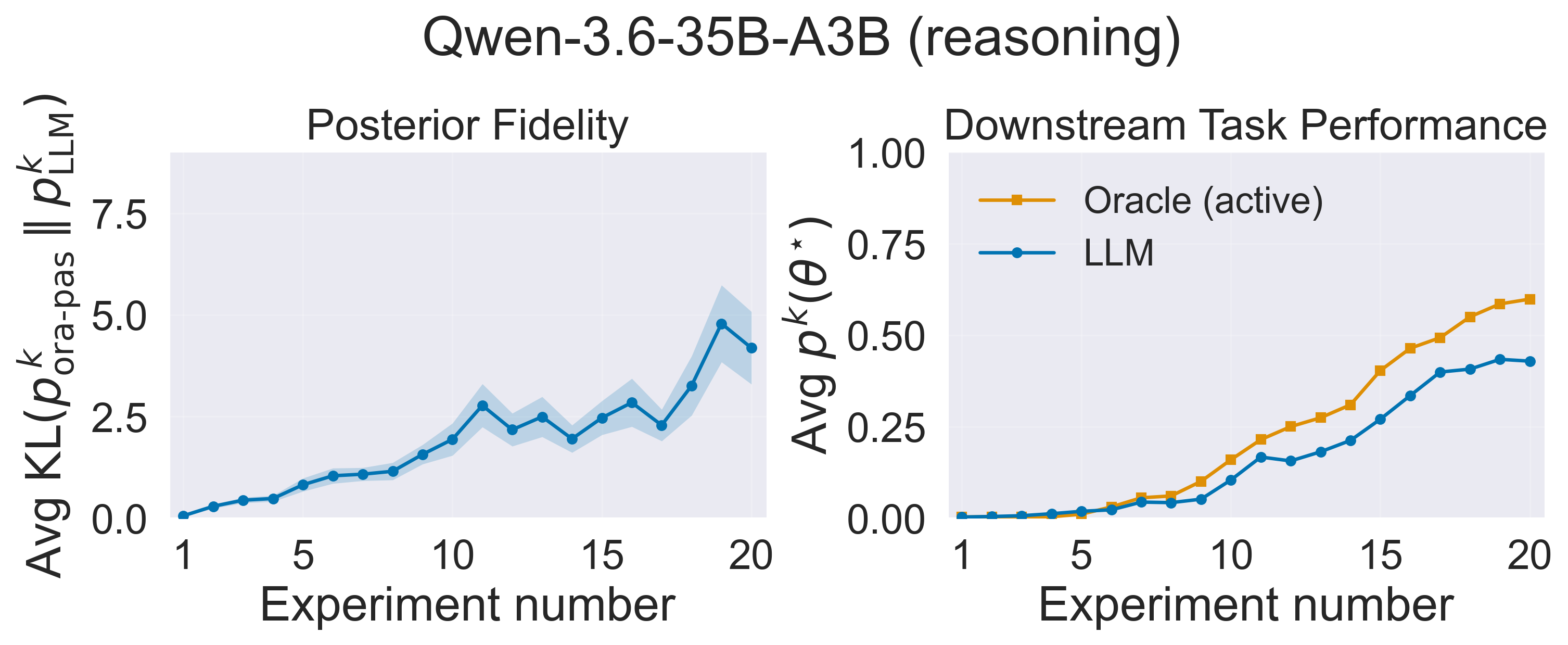}}
\end{minipage}
\end{center}

\newpage

\subsection{ProteinWorld}
\label{app:ProteinWorld_detailed_results}
\begin{center}
\captionof{figure}{Detailed results of posterior fidelity and downstream task performance on \texttt{ProteinWorld - simple} for all analyzed LLMs, as well as on of Gemini-2.5-Pro on \texttt{ProteinWorld - hard}.}

\noindent
\begin{minipage}[t]{0.495\textwidth}
    \centering
    \includegraphics[width=\linewidth]
    {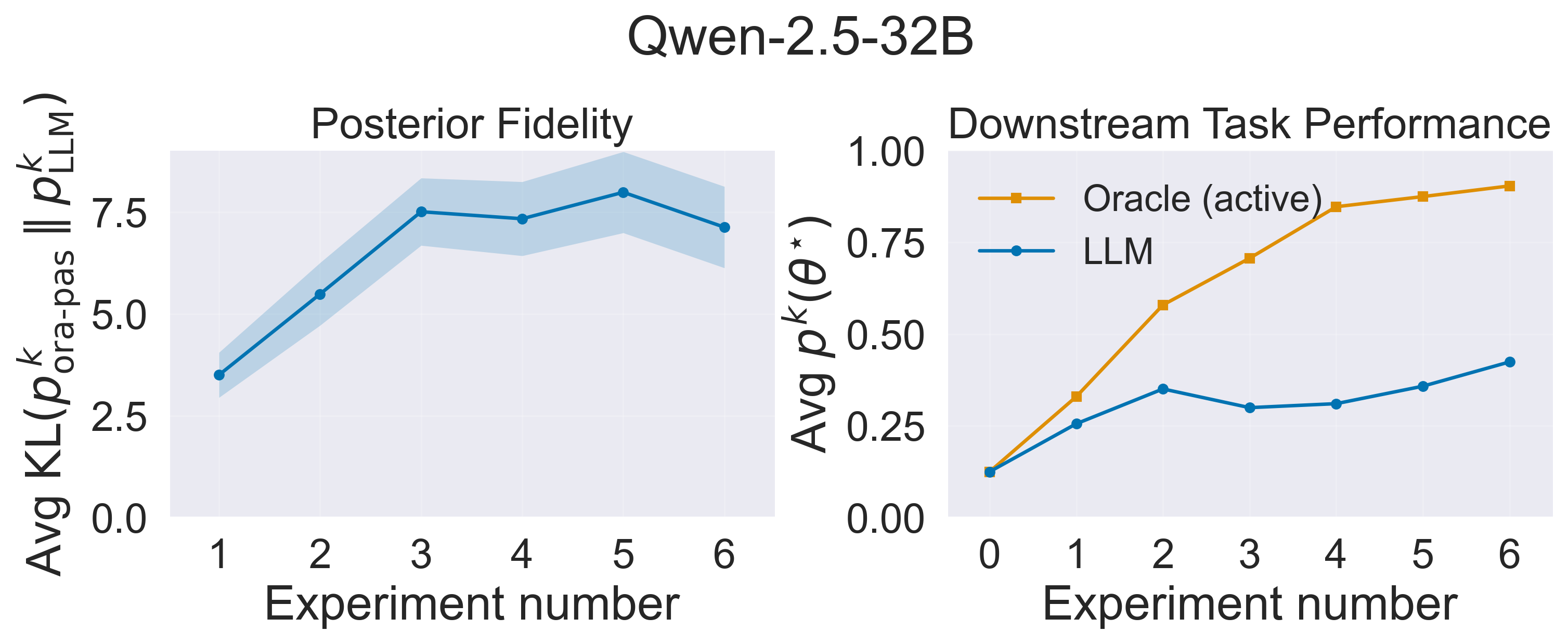}
\end{minipage}%
\hfill
\begin{minipage}[t]{0.495\textwidth}
    \centering
    \includegraphics[width=\linewidth]
    {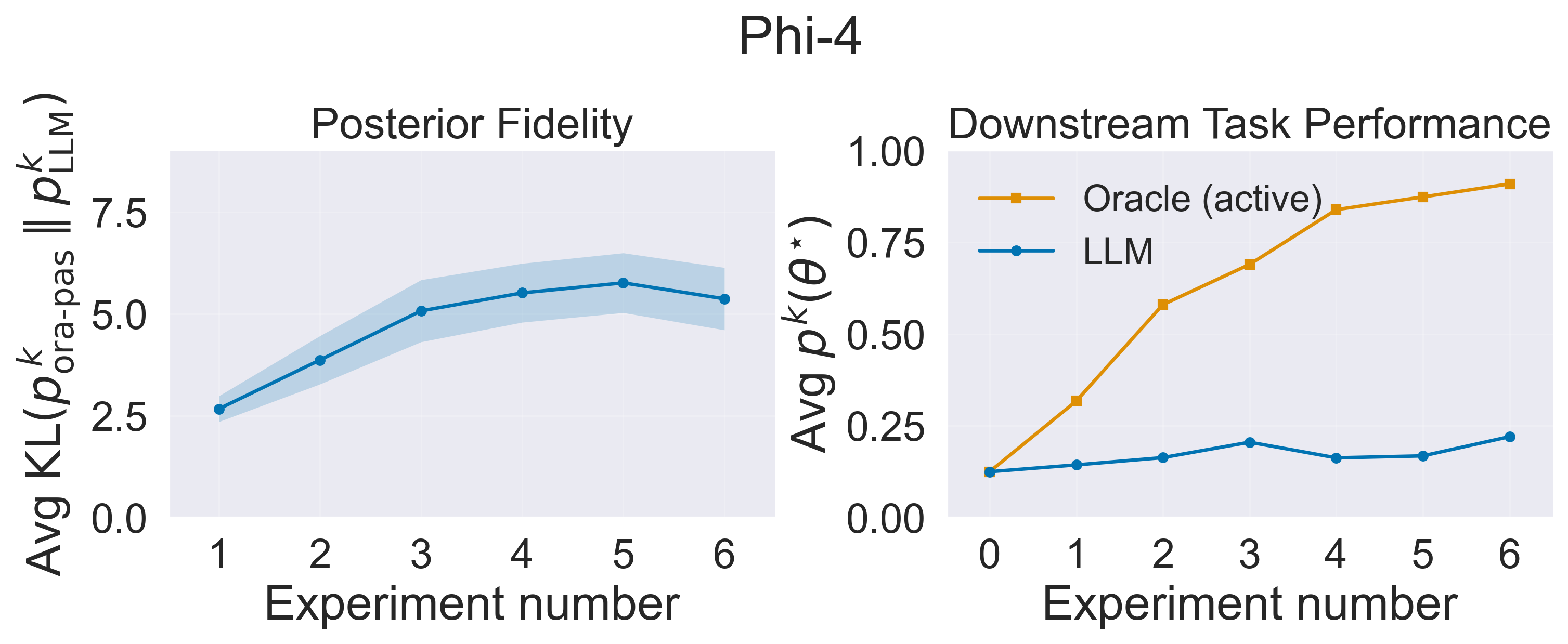}
\end{minipage}

\vspace{0.2em}

\noindent
\begin{minipage}[t]{0.495\textwidth}
    \centering
    \includegraphics[width=\linewidth]
    {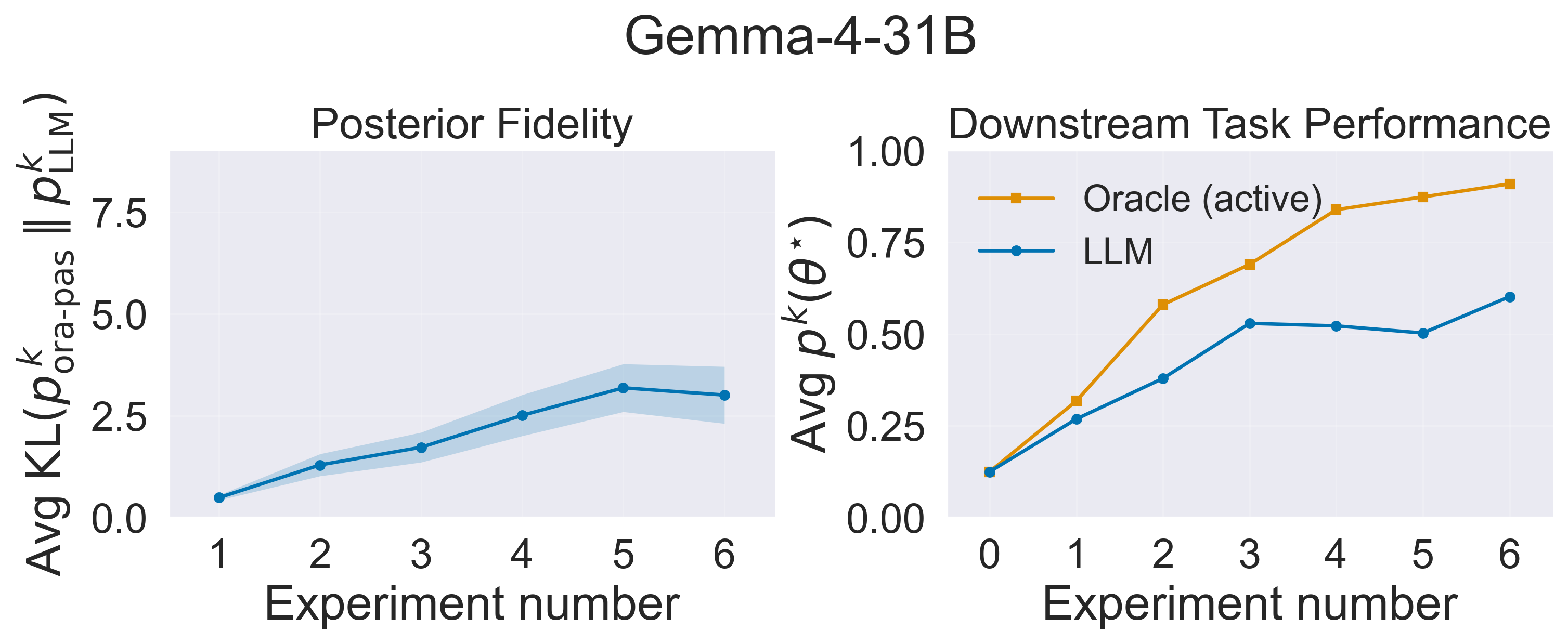}
\end{minipage}%
\hfill
\begin{minipage}[t]{0.495\textwidth}
    \centering
    \includegraphics[width=\linewidth]
    {\detokenize{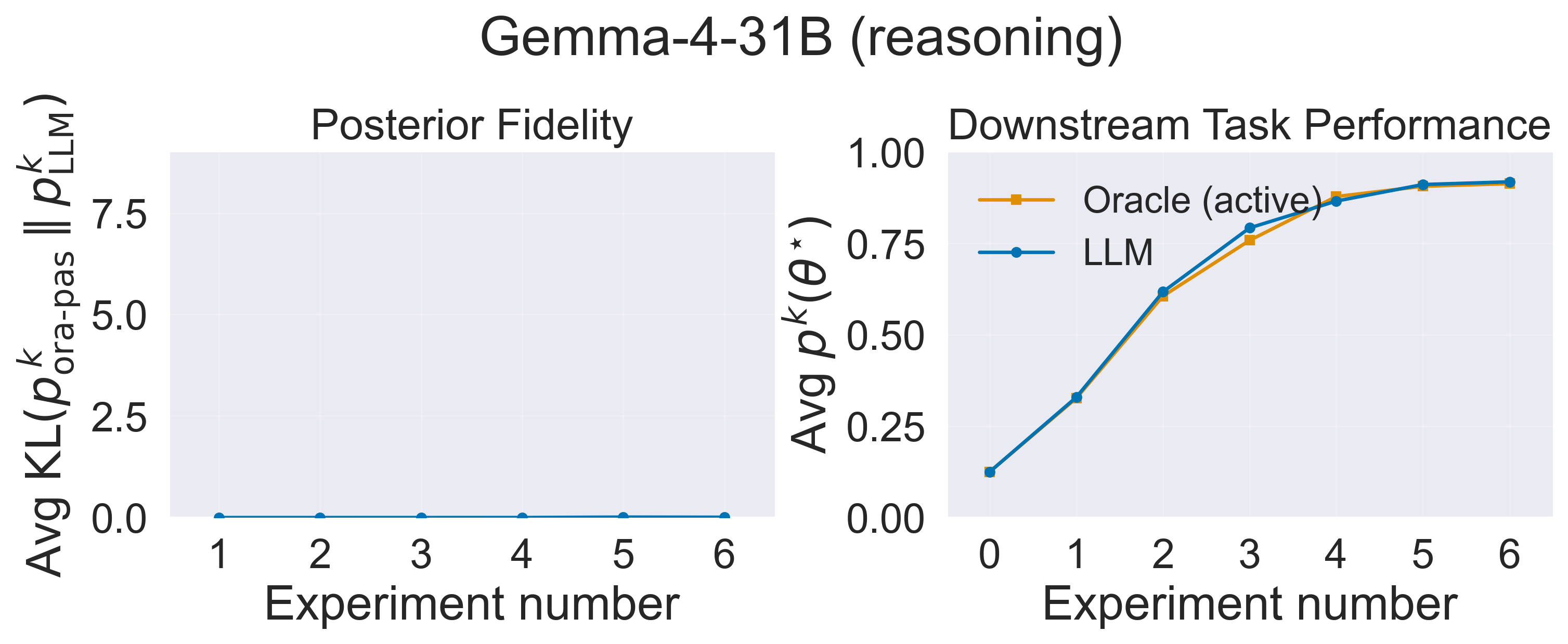}}
\end{minipage}

\vspace{0.2em}

\noindent
\begin{minipage}[t]{0.495\textwidth}
    \centering
    \includegraphics[width=\linewidth]
    {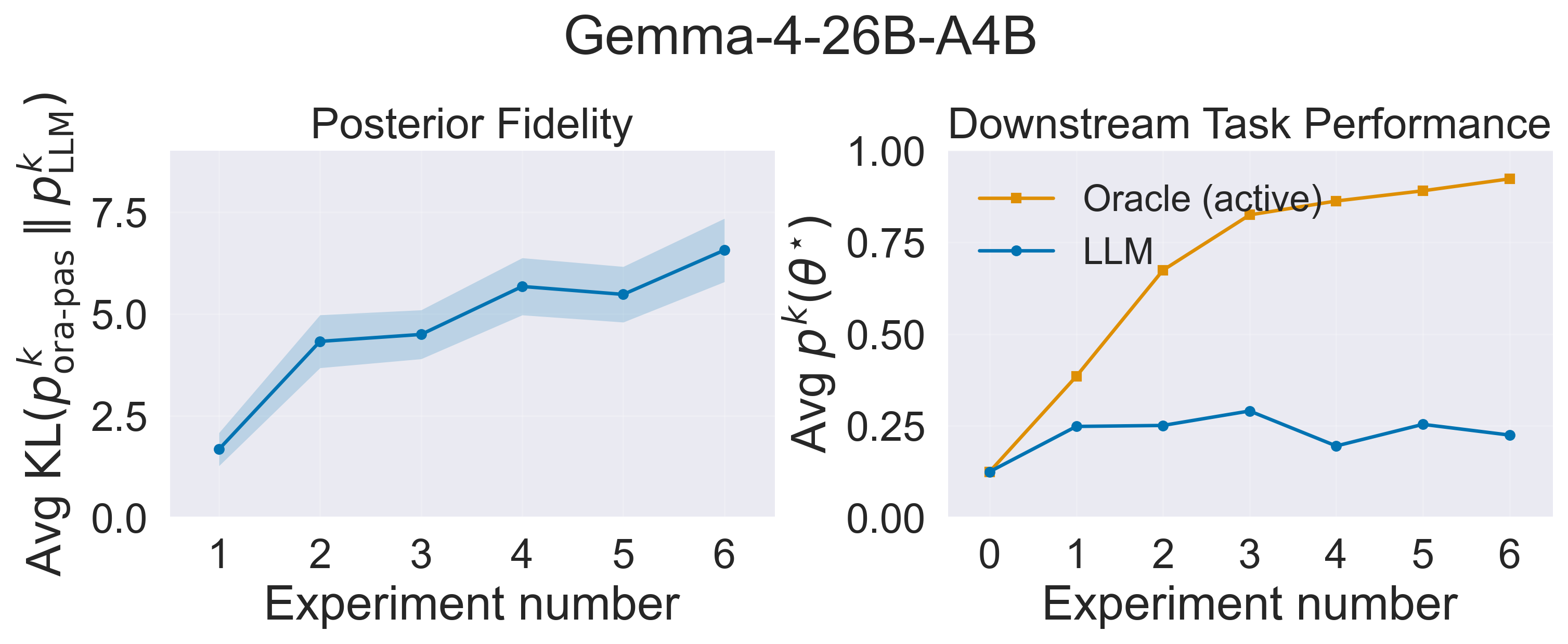}
\end{minipage}%
\hfill
\begin{minipage}[t]{0.495\textwidth}
    \centering
    \includegraphics[width=\linewidth]
    {\detokenize{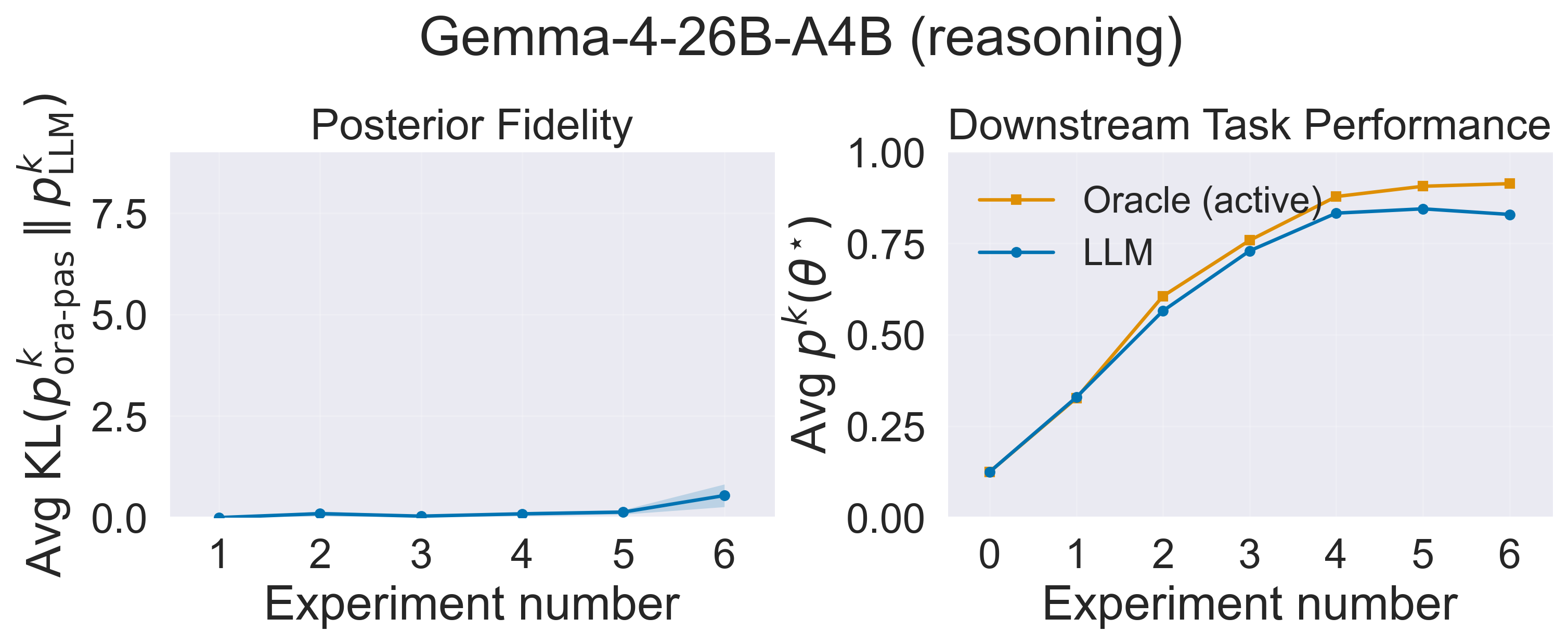}}
\end{minipage}

\vspace{0.2em}

\noindent
\begin{minipage}[t]{0.495\textwidth}
    \centering
    \includegraphics[width=\linewidth]
    {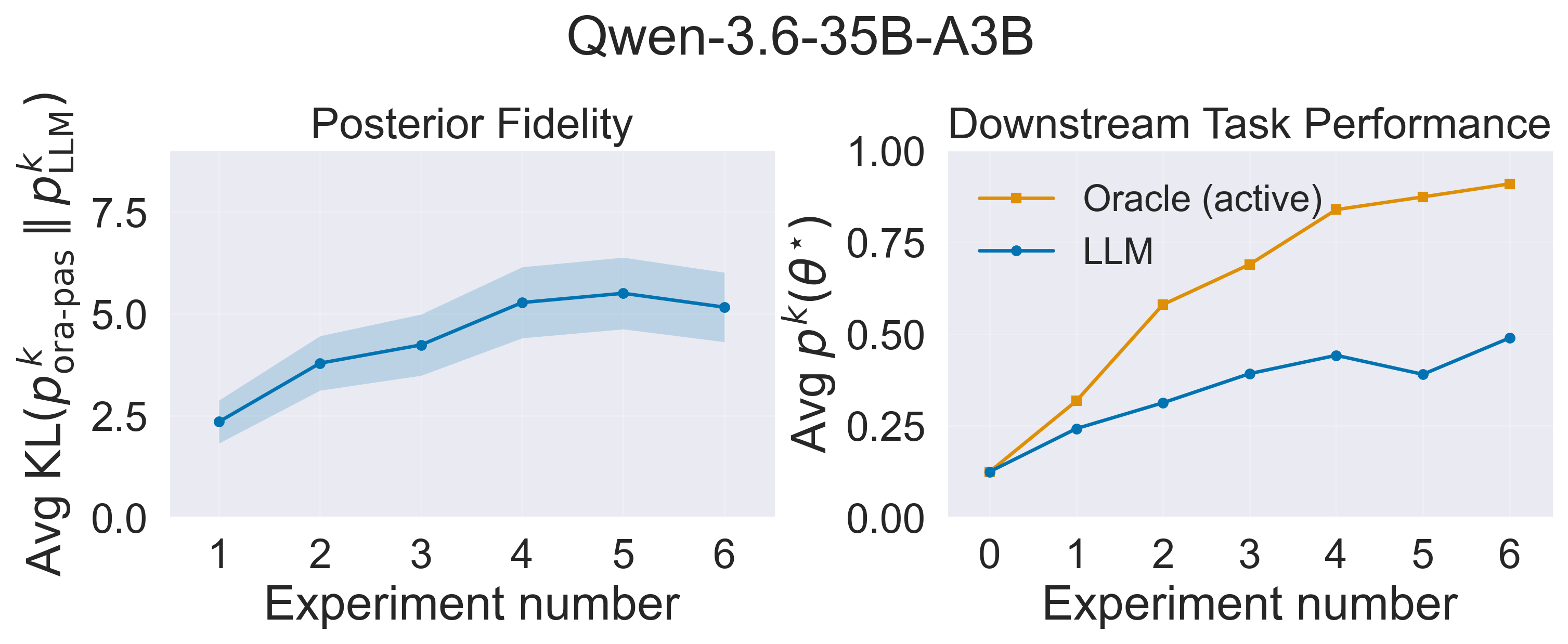}
\end{minipage}%
\hfill
\begin{minipage}[t]{0.495\textwidth}
    \centering
    \includegraphics[width=\linewidth]
    {\detokenize{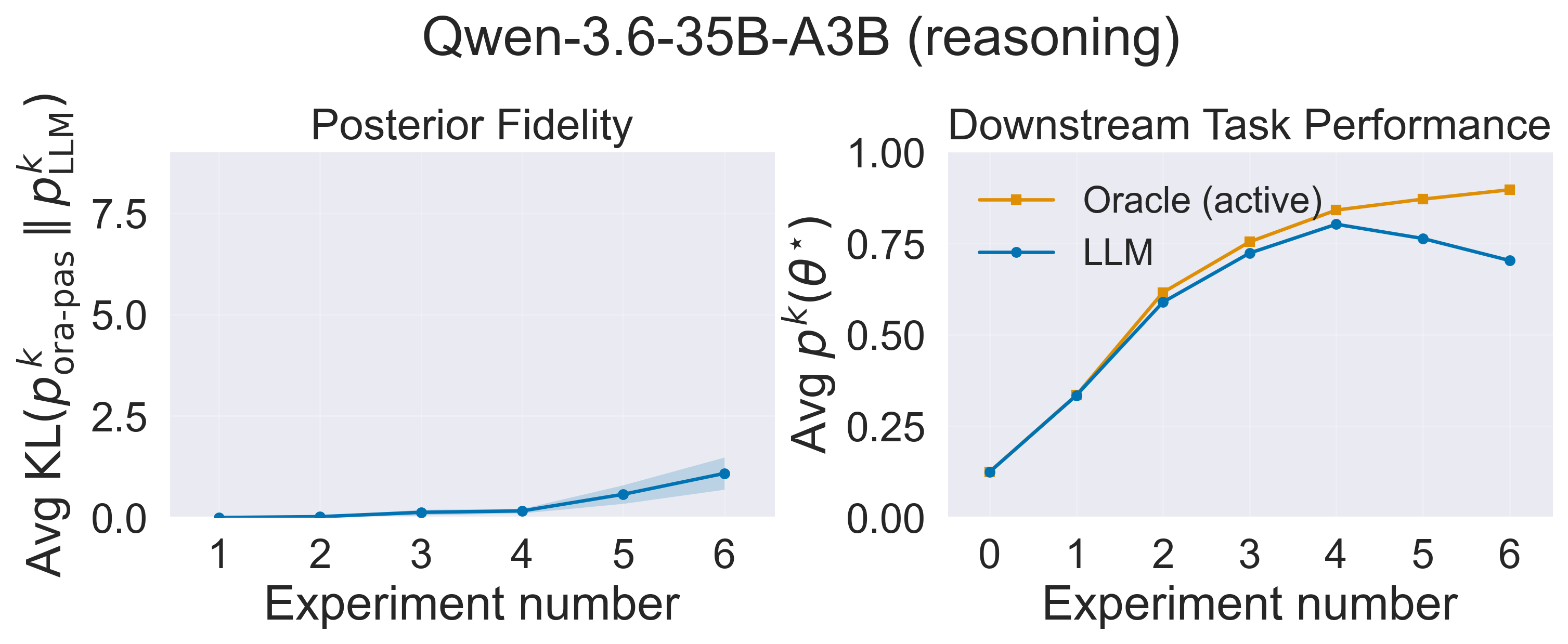}}
\end{minipage}

\vspace{0.2em}

\noindent
\begin{minipage}[t]{0.495\textwidth}
    \centering
    \includegraphics[width=\linewidth]
    {\detokenize{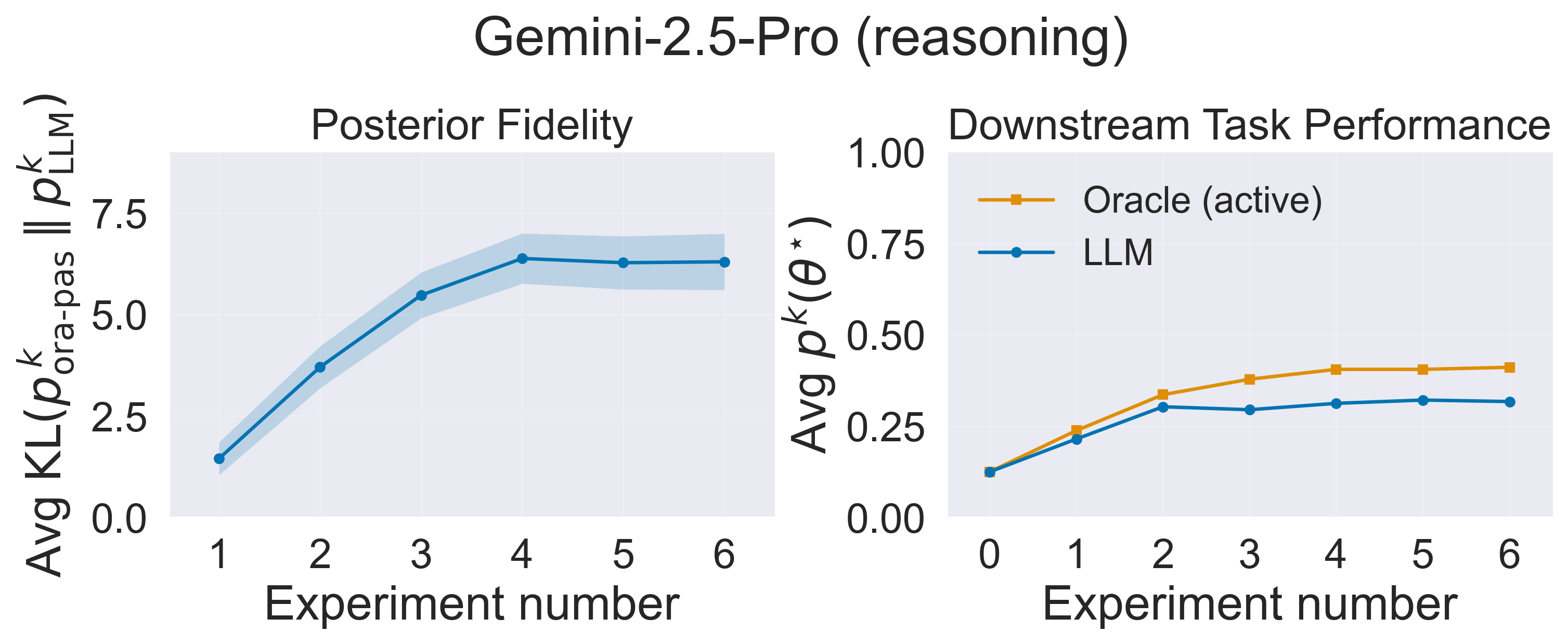}}
\end{minipage}%
\hfill
\begin{minipage}[t]{0.495\textwidth}
    \centering
    \includegraphics[width=\linewidth]
    {\detokenize{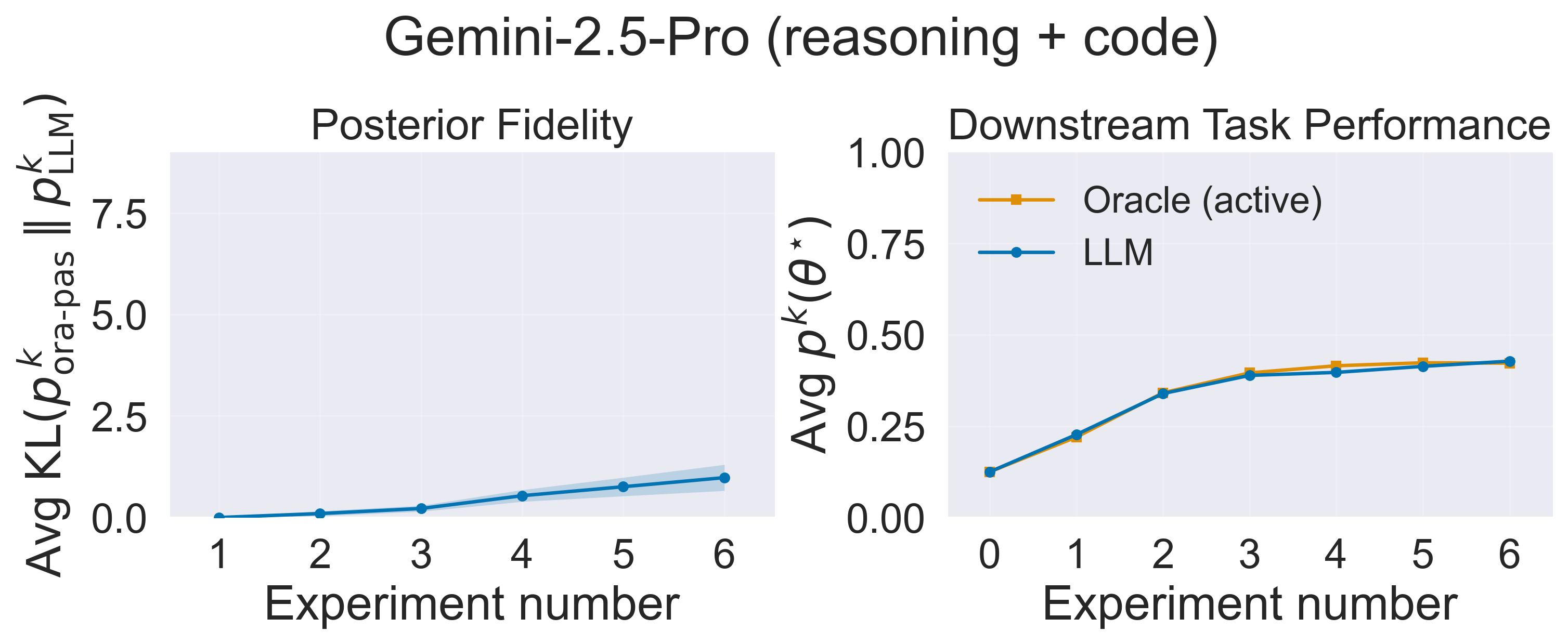}}
\end{minipage}
\end{center}

\newpage 

\section{Implementation Details}
\subsection{Dutch Books}
\subsubsection{Considered Prompts}\label{app:DB_scenarios}

\paragraph{Normalization Questions}
For each pair, the model is asked for the probability of the positive formulation and of its negation.

\begin{enumerate}
\item Positive: ``Will a Democrat win the 2028 US presidential election?'' \\
Negation: ``Will a candidate who is NOT a Democrat win the 2028 US presidential election?''
\item Positive: ``Will the People's Republic of China blockade or quarantine Taiwan before September 1, 2026?'' \\
Negation: ``Will the People's Republic of China NOT blockade or quarantine Taiwan before September 1, 2026?''
\item Positive: ``Will the United States attack Cuba before 2027?'' \\
Negation: ``Will the United States NOT attack Cuba before 2027?''
\item Positive: ``Will the Cuban government lose power before 2027?'' \\
Negation: ``Will the Cuban government still hold power on January 1, 2027?''
\item Positive: ``Will Elon Musk be removed as the Chief Executive Officer of Tesla before 2027?'' \\
Negation: ``Will Elon Musk still be the Chief Executive Officer of Tesla on January 1, 2027?''
\item Positive: ``Will Tesla discontinue the Cybertruck before January 1, 2027?'' \\
Negation: ``Will Tesla NOT discontinue the Cybertruck before January 1, 2027?''
\item Positive: ``Will Benjamin Netanyahu cease to be Prime Minister of Israel during 2026?'' \\
Negation: ``Will Benjamin Netanyahu remain Prime Minister of Israel throughout all of 2026?''
\item Positive: ``Will Keir Starmer cease to be Prime Minister of the UK during 2026?'' \\
Negation: ``Will Keir Starmer remain Prime Minister of the UK throughout all of 2026?''
\item Positive: ``Will Mary Peltola win the 2026 US Senate election in Alaska?'' \\
Negation: ``Will a candidate who is NOT Mary Peltola win the 2026 US Senate election in Alaska?''
\item Positive: ``Will Bitcoin trade above \$120,000 before August 1, 2026?'' \\
Negation: ``Will Bitcoin remain at or
below \$120,000 through August 1, 2026?''
\end{enumerate}

\paragraph{Monotonicity scenarios}
For each series, the model is asked for the probability of the templated question at each of the six listed threshold
values $x$, in increasing order.

\begin{enumerate}
\item ``Will Russia control at least $x$\% of Donbass at the end of October 2026?'', $x \in \{70, 75, 80, 85, 90, 95\}$

\item ``Will Nvidia's stock price close below \$$x$ on any day in 2026?'', $x \in \{150, 140, 130, 120, 110, 100\}$

\item ``Will the S\&P 500 decline by more than $x$\% in any 5-trading-day period before 2027?'', \\ $x \in \{5, 7, 10, 12, 15, 20\}$

\item ``Will the S\&P CoreLogic Case-Shiller U.S. National Home Price Index exceed $x$ by July 2027?'', $x \in \{320, 330, 340, 345, 350, 355\}$

\item ``Will Bitcoin's price exceed \$$x$ at any point before 2027?'', \\$x \in \{80000, 90000, 100000, 110000, 115000, 120000\}$

\item ``Will Anthropic's valuation reach at least \$$x$ trillion by December 31, 2026?'', \\$x \in \{1.0, 1.25, 1.5, 1.6, 1.7, 1.75\}$

\item ``Will total MLB regular-season attendance in 2026 exceed $x$ million?'', \\$x \in \{69, 70, 71, 71.5, 72, 72.5\}$

\item ``Will China's official youth unemployment rate fall below $x$\% in any monthly report released by 31 July 2026?'', $x \in \{18, 17, 16, 15.5, 15, 14\}$

\item ``Will Reform UK's poll average fall below $x$\% for at least 14 consecutive days in 2026?'', \\$x \in \{24, 22, 21, 20, 18, 15\}$

\item ``Will Spider-Man: Beyond the Spider-Verse get a $x$\% or higher on Rotten Tomatoes?'', $x \in \{90, 92, 93, 95, 96, 98\}$
\end{enumerate}                                                                                                     
  
\paragraph{Extensionality scenarios}
For each pair, the model is asked for the probability of two semantically equivalent formulations of the same event.

\begin{enumerate}
\item Q1: ``Will there be at least 1,000 deaths due to direct conflict between India and Pakistan before 2027?'' \\
Q2: ``Will the death toll from direct conflict between India and Pakistan reach four figures before 2027?''
\item Q1: ``Will federal courts block the USPS mail-in ballot restrictions before the 2028 US presidential election?'' \\
Q2: ``Will federal courts strike down the USPS mail-in ballot restrictions before the 2028 US presidential election?''
\item Q1: ``Will Nvidia's stock price close below \$100 on any day in 2026?'' \\
Q2: ``Will Nvidia stock dip under the \$100 mark at least once during 2026?''
\item Q1: ``Will the US stock market crash by more than 10\% in any 5-day trading period before 2027?'' \\
Q2: ``Will the S\&P 500 fall by over 10\% within some single 5-trading-day stretch before 2027?''
\item Q1: ``Will SpaceX successfully refuel a Starship in orbit during 2026?'' \\
Q2: ``Will SpaceX complete an in-orbit propellant transfer between two Starships at some point in 2026?''
\item Q1: ``Will Elon Musk become a trillionaire before July 2026?'' \\
Q2: ``Will Elon Musk's net worth cross the \$1 trillion mark before July 2026?''
\item Q1: ``Will the market capitalization for USDT \& USDC stablecoins surpass \$300 billion USD by June 30, 2026?'' \\
Q2: ``Will the combined market cap of USDT and USDC exceed \$300 billion before July 1, 2026?''
\item Q1: ``Will China's year-over-year GDP growth rate for Q2 2026 equal or exceed 5.0\%, according to the National Bureau of Statistics of China?'' \\
Q2: ``Will China's official year-over-year GDP growth for Q2 2026 come in at 5.0\% or higher?''
\item Q1: ``Will Waymo have a service area of 1,200 square miles before July 1, 2026?'' \\
Q2: ``Will Waymo's robotaxi service area expand to 1,200 square miles before July 2026?''
\item Q1: ``Will a hurricane make landfall in the continental United States before September 1, 2026?'' \\
Q2: ``Will a hurricane make landfall somewhere along the continental US coast before September 2026?''
\end{enumerate}

\subsubsection{Prompt setup}
The base event in each scenario is sourced verbatim or near-verbatim from ForecastBench; with complements, threshold-shifted variants, and alternate reformulations written by us as close as possible to the original style. 

All probability judgments were elicited using the same prompts. For each queried event, the model received the following instructions, where \texttt{{\{question\}}} was replaced by the corresponding ForecastBench question or logically related reformulation:

\begin{quote}
\texttt{System: You are an expert predictor of future events.} \\

\texttt{User: With what probability is the answer to the following question "Yes"?} \\

\texttt{{\{question\}}} \\

\texttt{Reply with exactly one decimal number between 0 and 1, which represents the probability.}
\end{quote}

Each probability was elicited using a temperature of $\tau=0.7$. Model responses were parsed as scalar probabilities in $[0,1]$. When eliciting probabilities in the same chat, the previous elicitation and response were simply kept in context as one user and one assistant message, and the same user message asking for a probability judgment of the new question was appended as final message.

\subsection{20 Questions}
Our implementation of \texttt{20Q} is adapted from \citet{bedllm}, formalized and adjusted to fit in our framework of Bayesian inference evaluation.

\subsubsection{Dynamic Hypothesis set $\Theta_k$}
Starting from an initial set $\Theta_0 \subset \Theta$, we construct $\Theta_k$ iteratively via $\Theta_k = f(\Theta_{k-1}; h_{1:k}) \cup f(\Theta_{\text{new}}; h_{1:k})$. Here, $\Theta_{\text{new}} \subset \Theta$ denotes a set of novel candidate beliefs generated conditionally on the history $h_{1:k}$, and $f(\cdot)$ is a filter function designed to prune invalid hypotheses. This dynamic expansion introduces a necessary deviation from strict Bayesian updating. Under standard Bayesian mechanics following Eq.~\ref{eq:sequential_bayes_update}, a hypothesis $\theta \notin \Theta_{k-1}$ inherently carries a prior probability $p_{k-1}(\theta) = 0$, which dictates that $p_k(\theta) = 0$ and forces the posterior space to strictly shrink ($\Theta_k \subseteq \Theta_{k-1}$). By deliberately injecting $\Theta_{\text{new}}$, we break this monotonic shrinking. This intervention is crucial; without it, the model could never evaluate beliefs omitted from the initial subset $\Theta_0$. While strict Bayesian adherence would require initializing $\Theta_0 = \Theta$, doing so is impossible, as the LLM cannot explicitly enumerate probabilities for every conceivable target.

The prompts used to generate $\Theta_0$ are these:
\begin{quote}
\texttt{System: You are playing a game of 20 Questions trying to guess an animal. To start the game, generate up to 50 candidate animals which might be the answer. Each candidate must be a single, existing animal (e.g., "Scarlet macaw", "White Shark", "Rhinoceros beetle"). List each animal on its own line - no numbering, punctuation, or extra text. Generate answers in English only. Produce a varied set by diversifying along different kinds of features. Don't count out more obscure animals. Do not repeat any animal. Return only the list of animals.} \\

\texttt{User: Let's start the game of 20 Questions. Generate a diverse set of animals, at least 15.}

\end{quote}

Similarly, here are the prompts to generate $\Theta_\text{new}$
\begin{quote}
\texttt{System: You are playing a game of 20 Questions trying to guess an animal. When asked to, using all of the questions and answers, generate up to 50 candidate animals that satisfy every clue, aiming for at least 15. Each candidate must be an existing animal (e.g., "Scarlet macaw", "White Shark", "Rhinoceros beetle"). List each animal on its own line - no numbering, punctuation, or extra text. Generate answers in English only. Produce a varied set by identifying features not implied by the clues and diversifying along them. Don't count out more obscure animals. Do not repeat any animal. Return only the list of animals.} \\

\texttt{Assistant: $\xi_1$} \\

\texttt{User: $y_1$} \\

\texttt{$\vdots$} \\

\texttt{Assistant: $\xi_k$} \\

\texttt{User: $y_k$} \\

\texttt{User: Now generate candidate animals fitting the questions and answers. Only generate animals that really fit ALL questions and answers. In your final response, don't include any commentary such as "this might be valid"; return only the animal names.}
\end{quote}

Generation happens with temperature $\tau = 1.3$ to encourage answer diversity. To prevent bad LLM responses from muddying the hypothesis set, all beliefs with more than $30$ characters are filtered out. Additionally, only the $100$ beliefs with highest probability are retained. Both of these filters are rarely needed and don't affect the final performance.

\subsubsection{LLM posterior}\label{app:20Q_LLM_posterior}
For each belief $\theta \in \Theta_k$, the LLM evaluates its compatibility with the experimental history $h_{1:k}$ by assigning a probability of it fitting to all pervious questions and answers:

\begin{quote}
\texttt{System: You are playing a game of 20 Questions trying to guess an animal. After asking Yes/No questions and receiving answers, you will be prompted to judge whether a given animal fits the questions and answers.} \\

\texttt{Assistant: $\xi_1$} \\

\texttt{User: $y_1$} \\

\texttt{$\vdots$} \\

\texttt{Assistant: $\xi_k$} \\

\texttt{User: $y_k$}
\end{quote}

Different prompts are used depending on whether probabilities are extracted from the model's token log-probabilities (non-reasoning models) or from verbalized probabilities (reasoning models).

\begin{quote}
\textbf{Token log-probabilities:} \\

\texttt{User: Does this animal} \\

\texttt{\{$\theta$\}} \\

\texttt{fit all the previous questions and answers? Reply exactly "Yes" or "No", depending on which best describes whether the animal fits. Consider every single question-answer pair and answer "Yes" if the animal fits all of them.}
\end{quote}

\begin{quote}
\textbf{Verbalized probabilities:} \\

\texttt{User: Does this animal: '\{$\theta$\}' fit all the previous questions and answers? Reply in the format:} \\

\texttt{Yes: <prob1>, No: <prob2>} \\

\texttt{where the probabilities are decimal numbers. Consider every single question-answer pair and give positive probability to "Yes" only if the animal fits all answers. Reply exactly in the specified format, and nothing else.}
\end{quote}

When using token log-probabilties, they are normalized so that the probabilities for "Yes" and "No" sum to 1. When using verbalized probabilities, if the LLM does not produce an output in the expected format, a default of \texttt{Yes: 0.5, No: 0.5} is used. This happens rarely (fewer than 1 in a 1000 times), and therefore does not affect performance. The temperature used is $\tau = 0.7$.

The compatibility score we then normalize across $\Theta_k$ is the probability the LLM assigns to answer "Yes" for each $\theta \in \Theta_k$. The filtering function $f(\cdot)$ filters out any beliefs whose compatibility score is lower than $0.2$.

\subsubsection{Oracle posterior}
We extract the LLM-based probabilities $p(y \mid \theta, \xi)$ using the following prompts, depending on whether answers are extracted from the model's token log-probabilities (non-reasoning models) or from verbalized probabilities (reasoning models):

\begin{quote}
\textbf{Token log-probabilities:} \\

\texttt{System: You are playing the answerer in a game of 20 Questions. Your chosen entity is:} \\

\texttt{\{$\theta$\}} \\

\texttt{When asked a question, you must reply exactly "Yes" or "No", depending on whether your chosen entity fulfills the question. You must not add any explanation, extra text, or anything else.} \\

\texttt{User: \{$\xi$\}}
\end{quote}

\begin{quote}
\textbf{Verbalized probabilities:} \\

\texttt{System: You are playing the answerer in a game of 20 Questions. Your chosen entity is:} \\

\texttt{\{$\theta$\}} \\

\texttt{When asked a Yes/No question, you must give probabilities to the answers "Yes" or "No", depending on how your chosen entity fits the question. Your answer must be in the format:} \\

\texttt{Yes: <prob1>, No: <prob2>} \\

\texttt{where the probabilities are decimal numbers. You must not add any explanation, extra text, or anything else.} \\

\texttt{User: \{$\xi$\}}
\end{quote}

Generation temperature, token log-probability normalization and defaults for verbalized probabilities work the same way as in App.~\ref{app:20Q_LLM_posterior}. Beliefs are filtered out if $p(y_i \mid \theta, \xi_i) \leq 0.2$ for any $i \in \{1,\dots,k\}$. 

\subsubsection{Experiment selection}
Due to the complex structure of $\Xi$, the EIG in \texttt{20Q} can't be maximised over all experiments, which is why we evaluate only a subset of fifteen candidate questions $\Xi_{k+1} \subset \Xi$, which is generated by the LLM conditioned on the current active hypothesis set $\Theta_{k}$. The prompt used is this:

\begin{quote}
\texttt{System: You are playing a game of 20 Questions to guess an animal. When asked, you must produce Yes/No questions. You must not add any explanation or extra text; only return the strict format requested by the user.} \\

\texttt{Assistant: $\xi_1$} \\

\texttt{User: $y_1$} \\

\texttt{$\vdots$} \\

\texttt{Assistant: $\xi_k$} \\

\texttt{User: $y_k$} \\

\texttt{User: Using the beliefs list: \{$\Theta_k$\} and all previous questions and answers, generate up to 15 candidate Yes/No questions that split the list into two roughly equal parts. Each question should be phrased so the answer is Yes or No. Do not repeat questions. List each question on its own line - no numbering, punctuation, or extra text.}
\end{quote}

Again, a temperature of $\tau = 1.3$ is used to encourage answer diversity.

Due to $\Theta$ being too large to calculate entropy over directly, we calculate the EIG of each experiment using the reduction in entropy over experimental outcomes. We therefore need to simulate the current distribution $p(\theta)$, both to calculate the first entropy term and to calculate the expected value already existing in the formula. We obtain this simulation $\Theta_k^\text{EIG}$ by sampling uniformly from $\Theta_k$, with $|\Theta_k^\text{EIG}| = \operatorname{min}\{|\Theta_k|, 15\}$.

An exception to this whole procedure is the case $|\Theta_k| \leq 2$, where the LLM directly checks if it is one of the two beliefs, since that question always maximizes EIG, using the question: $\xi_{k+1}=$ \texttt{Is it $\Theta_k\left[0\right]$?}

\subsubsection{Experiment execution}

Once experiment $\xi_{k+1}$ has been chosen, the experimental outcome $y_k$ is generated using the following prompt with temperature $\tau = 0.7$:

\begin{quote}
\texttt{System: You are playing the answerer in a game of 20 Questions. Your chosen entity is:} \\

\texttt{\{$\theta^\star$\}} \\

\texttt{When asked a question, you must reply exactly "Yes" or "No", depending on whether your chosen entity fulfills the question. If the question correctly guesses your chosen entity, reply exactly "Correct!". You must not add any explanation, extra text, or anything else.} \\

\texttt{User: \{$\xi_{k+1}$\}}
\end{quote}

If the experimental result is \texttt{Correct!}, we set $p(\theta^\star)_j=1$ $\forall j \geq k+1$, and finish the current run without recording the KL on any further questions.

\subsubsection{List of animals used}\label{app:20Q_animallist}
Each of the following 40 animals is used as one instance for \texttt{20Q}:

\texttt{Coyote, Rhinoceros beetle, Wolverine, Saiga antelope, Yak, Kiwi, Zebra, Golden jackal, Horseshoe crab, Sand cat, European Badger, Blue dragon sea slug, Cassowary, Electric eel, Flying fox, Giraffe, Hammerhead shark, Vicuña, Ibex, Aye-aye, Scarlet macaw, Peregrine falcon, Wildebeest, Capybara, Sea otter, Leafcutter ant, African elephant, Rosy boa, Quetzal, Snowy owl, Jerboa, Gharial, Albatross, Dhole, Kookaburra, Monarch butterfly, Uakari, Bottlenose dolphin, Platypus, Brown bear}

This is a subset of the 100 animals used in \citet{bedllm}.

\subsection{ProteinWorld}
\subsubsection{World generation}\label{app:prot_world_generation}
To generate a new world, fictional names are generated randomly using prompts like the following, sent to the answerer model at temperature $\tau=0.7$:

\begin{quote}
\texttt{System: You are a precise text generator. Follow formatting instructions exactly. Output only what is explicitly requested. No commentary or extra text. No explanations. No step-by-step reasoning. Just the answer.}

\texttt{User: Create exactly \{num\_protein\_classes\} fictional protein names.} \\
\texttt{Output format rules:} \\
\texttt{• Output one name per line} \\
\texttt{• Do not include any other text, labels, punctuation, headers, blank lines, or commentary.} \\
\texttt{Create plausible fictional protein names (no real protein names).}
\end{quote}

For \texttt{simple}, the attribute vector $\mathbf{a} = (a_1, \dots, a_J)$ of a given protein consists of independent Gaussian variables. The distribution of each component depends on the protein class:
\begin{equation*}
\begin{aligned}
a_i \mid \theta \sim \mathcal{N}(\mu_{\theta,i}, \sigma_{\theta,i}^2) \quad \forall i\in \{1,\dots,J\}, \qquad
\mu_{\theta,i} \sim \mathcal{N}(0,1), \
\sigma_{\theta,i}^2 \sim 0.1 + |\mathcal{N}(0,1)|.
\end{aligned}
\end{equation*}
Each experiment $\xi_k$ then reveals the value of one specified attribute $a_i$ exactly.

For \texttt{hard}, the world setup additionally specifies, for each $i \in \{2,\dots,J\}$, a parent set $\mathrm{pa}(i) \subseteq \{1,\dots,i-1\}$ that the $i$-th attribute causally depends on. Attribute $a_i$ is still Gaussian conditional on its parents and protein class, but its mean is now built from a per-class base level combined with a nonlinear transform of each parent's realized value:
\begin{equation*}
a_i \mid \mathrm{pa}(i), \theta \sim \mathcal{N}\bigl(\mu_{\theta,i}, \sigma_{\theta,i}^2\bigr), \qquad \sigma_{\theta,i} \sim 0.1 + |\mathcal{N}(0,1)|,
\end{equation*}
where $\mu_{\theta,i}$ is obtained by sequentially folding the base level with one contribution per parent, in parent order:
\begin{equation*}
\mu_{\theta,i} = \Bigl(\cdots\bigl(m_{\theta,i} \circ_1 c_1\bigr) \circ_2 c_2 \cdots\Bigr) \circ_{|\mathrm{pa}(i)|} c_{|\mathrm{pa}(i)|}, \qquad
c_j = w_j \, g_j(a_{p_j}),
\end{equation*}
with base level $m_{\theta,i}\sim\mathcal{N}(0,1)$ same as the \texttt{simple} case, weight $w_j \sim \mathcal{N}(0,1)$, transform non-linear transformation $g_j$ drawn uniformly from $\{x\mapsto x^2,\; x\mapsto\tanh(x)\}$, and combination operator $\circ_j$ drawn uniformly from $\{+,\times\}$, each independently per parent edge $a_j \to a_i$.

Experiment $\xi_k$ depends on a randomized set of multiple attributes: each attribute is included independently with probability $0.6$ (resampled until at least two are selected), combined into $\mu_k$ via the same fold-of-$\{+,\times\}$/$\{\mathrm{square},\tanh\}$ construction described above, and the observed outcome is drawn from one of three randomly chosen output families, with $\sigma\sim\mathrm{Uniform}(0.3,0.7)$:
\begin{equation*}
y_k \sim \mathcal{N}(\mu_k,\sigma^2), \qquad y_k \sim \mathrm{Uniform}(\mu_k-\sigma,\mu_k +\sigma), \qquad y_k \sim\mathrm{Bernoulli}(\mathrm{sigmoid}(\mu_k)).
\end{equation*}

\subsubsection{LLM posterior}
At iteration $k$, the LLM is given the entire world setup, including the exact distributions and parameters for $p(\mathbf{a} \mid \theta)$ and $p(y \mid \mathbf{a}, \xi)$, as well as a prior $p_0$ over $\Theta$ as system prompt:

\begin{quote}
{\ttfamily
System: You are an expert in the following fictional protein world: These are all existing proteins:

\{$\theta_1,\dots,\theta_M$\}.

Every protein has the following latent attributes:

\{description of $p(\mathbf{a}\mid\theta)$ for each attribute\}.

The following experiments to reveal information about the attributes exist:

\{description of $p(y\mid\mathbf{a},\xi)$ for each $\xi\in\Xi$\}.

The user has found a protein and wants to determine which one it is. Using your knowledge, you will give a current best estimate for the classification.
}
\end{quote}
The descriptions are presented as strings similar to the examples seen in App.~\ref{app:ProteinWorld_example_instances}.

It is then given the full history of experiments and results $(h_1,\dots,h_k)$, and asked to output the posterior probability distribution across $\Theta$, recorded as verbalized probabilities, sampled at temperature $\tau=1.0$:

\begin{quote}
{\ttfamily
User: You can assume that initially, this probability distribution over the protein classes held:

$p_0(\theta_1)$: $v_1$, $\dots$, $p_0(\theta_M)$: $v_M$.

I have ran the following experiments with the described outcomes:

$\xi_1$: $y_1$, $\dots$, $\xi_k$: $y_k$.

Using all of your information: What probability does my protein have to be in each class?

Reply in the format:
$\theta_1$: \texttt{<probability\_1>},
$\dots$,
$\theta_M$: \texttt{<probability\_M>}.

Make sure the result is a probability distribution: All probabilities have to be between 0 and 1, and they have to sum to 1.

Do not include any reasoning, numbering or extra text in the final output, only the requested format.
}
\end{quote}

In case that the output does not fit the required format, generation is attempted twice more using the same prompts and a different seed. Since \texttt{Phi-4} often did not manage to stick to the required format, the conversation was extended with \texttt{Assistant: $\theta_1$:}, to encourage following the correct format. Since all probabilities are still LLM-generated, this does not impact the analysis.

\subsubsection{Oracle posterior}\label{app:prot_world_oracle}
For \texttt{simple}, each attribute is Gaussian, meaning $\mathcal{A} = \mathbb{R}^n$, where n is the dimensionality of the attribute vector $\mathbf{a}$. Since experiment $\xi_k$ reveals $a_i$ exactly, Eq.~\ref{eq:likelihood_protworld} can be considerably simplified. 
\begin{align*}
&p(y_k \mid \theta, \xi_k) = \int_{\mathbb{R}^n} p(y_k \mid \mathbf{a}, \xi_k) p(\mathbf{a} \mid \theta) \, d\mathbf{a}
= \int_{\mathbb{R}^n} \delta(y_k - a_i) p(\mathbf{a} \mid \theta) \, d\mathbf{a} \\
= \int_{\mathbb{R}} \delta(y_k - &a_i) \left( \int_{\mathbb{R}^{n-1}} p(a_i, \mathbf{a}_{-i} \mid \theta) \, d\mathbf{a}_{-i} \right) da_i
= \int_{\mathbb{R}} \delta(y_k - a_i) p(a_i \mid \theta) \, da_i = p(a_i=y_k \mid \theta)
\end{align*}
Therefore, the likelihood has a closed-form solution: Since $a_i \mid \theta \sim \mathcal{N}(\mu_{\theta,i}, \sigma_{\theta,i}^2)$, 
\begin{equation*}
p(a_i = y_k \mid \theta) = \frac{1}{\sqrt{2\pi\sigma_{\theta,i}^2}} \exp\left( -\frac{(y_k - \mu_{\theta,i})^2}{2\sigma_{\theta,i}^2} \right).
\end{equation*}

The \texttt{hard} version purposefully prevents the existence of such a closed-form analytical solution: Each experiments depends on at least two Gaussian attributes, each of which has been transformed with one of the non-linear functions $\{x\mapsto x^2,\; x\mapsto\tanh(x)\}$. Accordingly, Eq.~\ref{eq:likelihood_protworld} is evaluated using MC estimation with $N=30000$. For each protein class $\theta$, we draw $N$ samples $\mathbf{a}^{(1)}_\theta, \dots, \mathbf{a}^{(N)}_\theta \sim p(\mathbf{a} \mid \theta)$ and calculate
\begin{equation*}
p(y_k \mid \theta, \xi_k) = \int_{\mathcal{A}} p(y_k \mid \mathbf{a}, \xi_k) p(\mathbf{a} \mid \theta) \, d\mathbf{a} \approx \frac{1}{N}\sum_{j=1}^{N} p(y_k \mid \mathbf{a}^{(j)}_\theta, \xi_k),
\end{equation*}
where $ p(y_k \mid \mathbf{a}^{(j)}_\theta, \xi_k)$ can be solved by evaluating the formula for $\mu_k$ as described in App.~\ref{app:prot_world_generation} using $\mathbf{a}^{(j)}_\theta$ as attribute values, and then using the appropriate likelihood function for Gaussian, uniform, or Bernoulli(sigmoid), depending on $\xi_k$.

\subsubsection{Experiment selection}
Due to its fixed and finite size, we can maximize EIG over the entire set $\Xi$. For this environment, the experimental result space $\mathcal{Y}_\xi$ is the one that could be complex, while $\Theta$ has a pre-defined, simple structure. Therefore, calculating the entropy over $\Theta$ is more tangible, meaning that we calculate the EIG of each experiment by using the reduction in entropy over hypotheses.

To calculate this reduction for each experiment, we use MC estimation of size $M=500$ for the integral over $p(y\mid \xi)$. We employ forward sampling of the joint distribution of $(\theta, y)$ by first simulating $\theta^{(j)}$ according to the current distribution and then $y^{(j)}$ using $\theta^{(j)}$:
\begin{equation*}
\theta^{(j)} \sim p(\theta), y^{(j)} \sim p(y \mid \theta^{(j)}, \xi) \quad\Longrightarrow\quad y^{(j)} \sim p(y\mid\xi)
\end{equation*}
This allows calculating the EIG in the following way:
\begin{align*}
\operatorname{EIG}(\xi) = \mathbb{E}_{y \sim p(y \mid \xi)} \!\left[ \mathrm{H}\!\left[p(\theta)\right] - \mathrm{H}\!\left[p(\theta \mid \xi, y)\right] \right] = \mathrm{H}\!\left[p(\theta)\right] - \frac{1}{M}\sum_{j=1}^{M}  \mathrm{H}\!\left[p(\theta \mid \xi, y^{(j)})\right] 
\end{align*}
To further progress, Bayes' rule can be used:
\begin{align*}
p(\theta \mid \xi, y) \propto p(y \mid \theta, \xi) p(\theta),
\end{align*}
where $ p(y \mid \theta, \xi)$ can be calculated/approximated as seen in App.~\ref{app:prot_world_oracle}.

\end{document}

%% file: math_commands.tex
\usepackage{amsmath,amsfonts,bm}

\def\eqref#1{equation~\ref{#1}}

\def\1{\bm{1}}

\DeclareMathAlphabet{\mathsfit}{\encodingdefault}{\sfdefault}{m}{sl}
\SetMathAlphabet{\mathsfit}{bold}{\encodingdefault}{\sfdefault}{bx}{n}

